%% file: main.tex
\documentclass[10pt,twocolumn,letterpaper]{article}

\usepackage[letterpaper,margin=0.75in]{geometry}
\usepackage[utf8]{inputenc}
\usepackage{mathptmx}
\usepackage{amsmath}
\usepackage{amssymb}
\usepackage{amsthm}
\usepackage{graphicx}
\usepackage{booktabs}
\usepackage[font=small]{caption}
\usepackage[ruled,linesnumbered,lined]{algorithm2e}
\usepackage[numbers]{natbib}
\usepackage[hyphens]{url}
\usepackage[hidelinks]{hyperref}

\newtheorem{proposition}{Proposition}

\newcommand{\bkv}{b_{kv}}
\newcommand{\bs}{b_s}
\newcommand{\Hkv}{H_{kv}}
\newcommand{\nhalf}{n_{1/2}}
\newcommand{\abar}{\bar\alpha}
\newcommand{\dbar}{\bar\delta}
\newcommand{\KV}{|\mathrm{KV}|}
\newcommand{\dnll}{\Delta\mathrm{NLL}}

\begin{document}

\twocolumn[{%
\begin{center}
\vspace*{-0.10in}
\hrule height 4pt
\vspace{18.06pt}
{\fontsize{17.28}{27.7}\selectfont\bfseries
Joint KV-Cache and Recurrent-State Quantization\\
in Hybrid Linear-Attention LLMs\par}
\vspace{18.44pt}
\hrule height 0.4pt
\vspace{22.0pt}
{\large\bfseries Minseok Shin\par}
\vspace{4pt}
Horcrux Technologies, Seoul, Republic of Korea\\
\url{https://www.horcruxtechnologies.com}\\
\texttt{horcruxtech.ai@gmail.com}\quad\texttt{stevenshin16@korea.ac.kr}
\par\vspace{27.5pt}
{\large\bfseries Abstract\par}
\vspace{0pt}
\end{center}
\begin{list}{}{\setlength{\leftmargin}{0.6in}\setlength{\rightmargin}{0.6in}\setlength{\topsep}{0pt}}
\item[]
Hybrid language models that interleave linear-attention layers with softmax attention keep two inference
memories: a KV cache that grows with the context and a fixed-size recurrent state. Post-training
quantization has studied each in isolation, but a serving engine must split one budget between them. We
present the first joint study of post-training, calibration-free quantization of both memories in Gated
DeltaNet hybrids (Qwen3.5, 0.8B to 9B), with a Mamba-2 hybrid (Falcon-H1) as control. The asymmetry is
large: 4-bit KV costs at most 0.03 nats, whereas a 4-bit state costs about one nat; in the Mamba-2 hybrid
the same 4-bit state costs nothing, pointing to the delta rule. The two errors are only weakly
super-additive, so the memories can be allocated almost independently. A noise-matched control shows that
round-to-nearest is idempotent, which keeps state error far below a worst-case bound, while per-layer
sensitivity is concentrated in the first layer rather than following the decay half-life. The limiting
memory is task dependent: retrieval is bounded by the KV cache and reasoning by the state, where an 8-bit
state lowers GSM8K accuracy from 0.600 to 0.345 unless error feedback is used. On Qwen3.5-4B, 4-bit KV
with a 6-bit state and error feedback uses 32\% of the bf16 cache for 0.04 nats and GSM8K 0.555, and
4-bit KV with an 8-bit state and error feedback is lossless at 40\%.

\vspace{0.08in}
\textbf{Keywords:} large language models; hybrid linear attention; Gated DeltaNet; KV-cache quantization;
recurrent-state quantization; post-training quantization; memory-efficient inference.
\end{list}
\vspace{0.25in}
}]

\input{sections/01_introduction}
\input{sections/02_related_work}
\input{sections/03_background}
\input{sections/04_method}
\input{sections/05_setup}
\input{sections/06_results}
\input{sections/07_analysis}
\input{sections/08_discussion}
\input{sections/09_conclusion}

\bibliographystyle{plain}
\bibliography{refs}

\clearpage
\onecolumn
\appendix
\input{sections/appendix}

\end{document}

%% file: sections/01_introduction.tex
\section{Introduction}
\label{sec:intro}

The memory that a language model keeps during generation used to be one thing: the key-value (KV) cache
of its attention layers, growing linearly with the context. In 2026 the dominant open-weight families no
longer look like that. Qwen3.5, Kimi Linear and Nemotron interleave a few softmax-attention layers with
many linear-attention layers, and the linear layers keep a fixed-size recurrent state instead of a
cache~\cite{kimi2025linear,qwen2026qwen35,yang2025gdn}. A hybrid model therefore carries two memories
with different shapes, different growth laws and, as we show, different responses to low-precision
storage.

\begin{table}[t]
\caption{Headline results on Qwen3.5-4B. Mem.: footprint of both caches at $T=512$ relative to bf16
(eqs.~\ref{eq:kv}--\ref{eq:state}). $\dnll$: 3 seeds $\times$ 20 WikiText-2 windows. GSM8K: 200
problems, thinking mode. NIAH: mean over 4k and 16k. -- = not evaluated.}
\label{tab:headline}
\setlength{\tabcolsep}{3pt}
\resizebox{0.995\linewidth}{!}{\input{tables/table1}}
\vspace{6pt}
\end{table}

Quantizing inference memory is the standard way to fit more context or more concurrent requests on a
device, and for the KV cache it is mature: 4-bit storage is nearly free and 2-bit storage is usable with
care~\cite{hooper2024kvquant,liu2024kivi}. For the recurrent state the picture is much newer and much
less forgiving: the first post-training studies, all from the second half of 2026, report that even 8-bit
storage can hurt and that 4-bit storage breaks the
model~\cite{zhang2026damp,pan2026leapquant,yao2026stepquant,jiang2026scale}. Each of these works, however, treats one
memory in isolation and leaves the other in full precision. Nobody has asked the question a serving
engine actually faces: given a total memory budget, how should the bits be split between the KV cache
and the recurrent state?

\begin{figure*}[t]
\centering
\includegraphics[width=\textwidth]{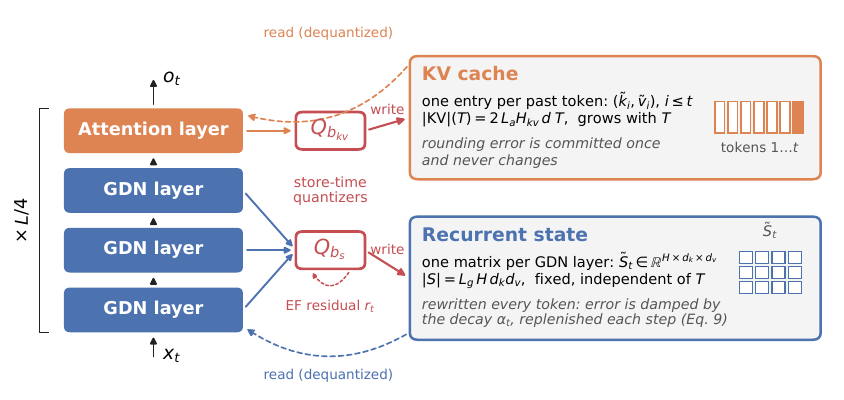}
\caption{Overview. One Qwen3.5 block (three Gated DeltaNet layers, one gated attention layer), its two
inference memories, and the two store-time quantization hooks. Error feedback (dashed) keeps a
full-precision residual beside the quantized state.}
\label{fig:overview}
\end{figure*}

This paper answers that question empirically and explains the answer analytically. Three properties of
hybrid models make it non-trivial. First, the two memories trade places as the context grows: below a
crossover length $T^\star$ (768 tokens for Qwen3.5-4B) the recurrent state is the larger memory, above it
the KV cache is. Second, their errors propagate differently. A rounded key is wrong once and stays wrong;
a rounded state is fed back into a recurrence, so its error is damped by the model's learned decay gates
but replenished at every token, converging to a steady state that depends on the layer. Third, the two
errors meet: attention layers read hidden states produced by quantized recurrences and vice versa, so
the loss under joint quantization need not be the sum of the losses under each alone.

We study these questions with store-time fake quantization, a post-training scheme that needs no
calibration data and leaves the weights untouched, on Qwen3.5 models from 0.8B to 9B parameters, and on
Falcon-H1 as a Mamba-2 based control, measuring teacher-forced decoding loss, mathematical reasoning in
thinking mode and long-context retrieval, with and without error feedback on the state. Our
contributions are:
\begin{itemize}
\item A joint bit-allocation study of KV cache and recurrent state under a shared memory budget, the
first to our knowledge, with a Pareto analysis across context length, model scale and task. The
asymmetry between the two memories is large and stable: 4-bit KV costs at most 0.03 nats at every scale,
4-bit state costs about one nat, and the state needs 8 bits to come within 0.02 nats of bf16.
\item An error-propagation model for the gated delta rule (Section~\ref{sec:errprop}) whose steady-state
bound explains the asymmetry, together with a test of its per-layer prediction that is \emph{not}
supported: at both 0.8B and 4B a single early layer dominates the summed single-layer sensitivity
(38\% and 80\%), in line with the first-layer concentration reported concurrently on smaller
models~\cite{jiang2026scale}, and the correlation with the decay half-life is absent at both sizes (Spearman $-0.08$
and $+0.13$). A noise-matched control explains why the bound is loose: round-to-nearest is idempotent, so
re-storing an already rounded state injects no new error, whereas the bound assumes fresh error at every
token.
\item A direct test of additivity between the two error sources. Joint quantization is weakly
super-additive at small scale (excess $+0.03$ to $+0.05$ nats at the 4-bit/4-bit corner on 0.8B and 2B
and on a second corpus) and additive at 9B, so the two memories can be allocated almost independently.
\item Evidence that the memory that limits accuracy depends on the task: the state for language
modelling, the KV cache for long-context retrieval (only 2-bit KV hurts, 4-bit state does not), and the
state for reasoning: on Qwen3.5-4B an 8-bit state, lossless on language modelling, lowers GSM8K accuracy
from 0.600 to 0.345 while 4-bit KV is lossless (0.595), and error feedback restores the 8-bit state to
bf16 accuracy (0.600) and a 6-bit state to 0.560.
\item A practical recipe with error feedback always on: on Qwen3.5-4B, 4-bit KV with a 6-bit state and
error feedback costs 0.04 nats at 32\% of the bf16 footprint at 512 tokens and keeps GSM8K accuracy at
0.555 against 0.600, not distinguishable from bf16 (Table~\ref{tab:headline}); 4-bit KV with an 8-bit
state is lossless at 40\% of the bf16 footprint on Qwen3.5-4B ($+0.003$ nats RTN, $-0.001$ with error
feedback), with retrieval at 1.00; at long context the KV cache dominates memory and the state bits
should be the last to go.
\end{itemize}
Code, configurations and raw results are released at \url{https://github.com/skytinstone/hybridquant}.

\paragraph{Organization.} Section~\ref{sec:related} positions the work; Section~\ref{sec:background}
fixes notation and derives the memory crossover; Section~\ref{sec:method} defines store-time
quantization, the error-propagation model with its steady-state bound
(Proposition~\ref{prop:steady}), error feedback, and the allocation problem together with the metrics
we report; Sections~\ref{sec:setup} to~\ref{sec:analysis} give the experimental setup, main results and
analysis; Section~\ref{sec:discussion} draws the practical recipe and states limitations. Proofs and full
result tables are in the appendix.

%% file: tables/table1.tex
\begin{tabular}{lrrrr}
\toprule
Setting & Mem.\ (\%) & $\Delta$NLL & GSM8K & NIAH \\
\midrule
bf16 & 100 & 0 & 0.600 & 1.00 \\
KV-only (KV4) & 70 & $+0.001$ & 0.595 & 1.00 \\
State-only (S8, EF) & 70 & $-0.000$ & 0.600 & 1.00 \\
Joint KV4 + S6, EF & 32 & $+0.044$ & 0.555 & 1.00 \\
Joint KV4 + S8 & 40 & $+0.003$ & -- & 1.00 \\
Joint KV4 + S8, EF & 40 & $-0.001$ & -- & 1.00 \\
\bottomrule
\end{tabular}

%% file: sections/02_related_work.tex
\section{Related Work}
\label{sec:related}

\paragraph{KV-cache quantization.} Compressing the KV cache of softmax attention is well studied. KIVI
quantizes keys per channel and values per token to 2 bits without tuning~\cite{liu2024kivi}; KVQuant
adds pre-RoPE key quantization, non-uniform codebooks and outlier isolation to reach very long
contexts~\cite{hooper2024kvquant}; GEAR pairs low-bit codes with a low-rank residual~\cite{kang2024gear};
rotation-based methods such as QuaRot and KVLinC flatten outliers before
quantization~\cite{ashkboos2024quarot,saxena2025kvlinc}. A second line targets reasoning models whose long
chains of thought accumulate KV error: PM-KVQ lowers precision progressively during
generation~\cite{liu2025pmkvq}, and MixKVQ allocates per-channel precision from query
relevance~\cite{zhang2025mixkvq}; an empirical study finds 4-bit KV close to lossless and 3-bit harmful
for reasoning~\cite{liu2025quanthurts}. All of these assume a pure attention model with a single,
token-indexed memory.

\paragraph{Recurrent-state quantization.} Hybrid models expose a second memory that these methods do not
touch. Post-training quantization of the recurrent state has been studied only since mid-2026. DAMP
reports that 8-bit state storage already degrades GDN and KDA models and proposes decay-aware mixed
precision across channels~\cite{zhang2026damp}; STEPQuant allocates bits by the lifetime of each state
unit~\cite{yao2026stepquant}; LeapQuant keeps a few high-precision compensator tokens next to an 8-bit
state~\cite{pan2026leapquant}; DASC selects which state units to checkpoint at all~\cite{yu2026dasc}.
A second wave refines this picture. Chen and He derive per-channel distortion weights from the
observability Gramian for calibration-free mixed precision~\cite{chen2026lowbit}; Li et al.\ trace error
accumulation to small updates that are lost in rounding and preserve them with a low-rank bypass and
directional rounding~\cite{li2026preserving}; Huang et al.\ allocate per-head bit-widths from gradient
sensitivity and keep recent writes in a higher-precision sliding window~\cite{huang2026gga}; Nazari et
al.\ combine value-axis scale grouping, decay-based mixed precision and stochastic rounding for 8-bit
states~\cite{nazari2026lores}; and Jiang identifies a feedback between the quantizer scale and the state
and reports that single-layer sensitivity on Qwen3.5-0.8B and 2B is concentrated in the first GDN
layer~\cite{jiang2026scale}, a concentration that our layer sweep also finds and extends to 4B
(Section~\ref{sec:layer-sensitivity}).
Related work on weight quantization shows that GDN projections survive 4-bit
NVFP4~\cite{kozyrev2026nvfp4}, and a study of recurrent-state write-back in a GRU model from a different
domain isolates the same freezing effect that coarse state storage produces~\cite{erbas2026writeback}.
To our knowledge, every one of these works quantizes the state with the KV cache held at full
precision. We quantize both memories together, under a shared budget, and ask how the bits should be
split between them.

\paragraph{Mixed precision and bit allocation.} Several KV methods allocate precision non-uniformly
inside the cache: QAQ keeps attention-sensitive tokens at higher precision~\cite{dong2024qaq}, KVTuner
searches a layer-wise mixed-precision schedule~\cite{li2025kvtuner}, and rate--distortion formulations
choose per-channel bit-widths under a budget~\cite{zuo2026ratequant}. On the state side, DAMP and
STEPQuant allocate bits across channels or units of a single state. Our allocation
problem~\eqref{eq:alloc} sits one level above these: it decides how much of the budget each memory
receives, and any intra-memory scheme can be applied inside the resulting split. We therefore use
uniform group quantization within each memory and leave the composition with intra-memory schemes to
future work.

\paragraph{Scope.} We study post-training, data-free quantization of inference memory with the weights
untouched. Quantization-aware training and distillation (e.g.\ QAD on
Qwen3.5-4B~\cite{kim2026qad}) and weight-only PTQ (GPTQ, AWQ~\cite{frantar2023gptq,lin2024awq}) are
complementary and out of scope; they change the model, whereas we change only what is stored between
decoding steps.

\paragraph{Hybrid architectures and serving.} Gated DeltaNet combines the delta
rule~\cite{schlag2021fwp,yang2024deltanet} with Mamba-2 style gating~\cite{dao2024mamba2,yang2025gdn} and
is the dominant mixer in Qwen3.5~\cite{qwen2026qwen35}; Kimi Linear uses a channel-wise decay
variant~\cite{kimi2025linear}. Serving engines store the state in bf16 regardless of KV precision, which
is why 8-bit KV storage yields no memory saving on hybrids at short context and $1.84\times$ rather than
$2\times$ at 32k tokens~\cite{vllm2026fp8kv}. A comment in the same discussion reports, from the
commenter's own runs on Qwen3.5-4B, that INT8 state storage lowers GSM8K accuracy from 81\% to 41\% and
that error feedback restores it to 81\%. Our controlled measurement agrees in direction and magnitude of
the repair (0.600 to 0.345, restored to 0.600 with error feedback; Section~\ref{sec:task}); our bf16
accuracy is lower because of the 4,096-token budget, which 48.5\% of bf16 generations reach.

%% file: sections/03_background.tex
\section{Background}
\label{sec:background}

\subsection{Hybrid linear-attention language models}
\label{sec:hybrid}

Recent open-weight LLM families replace most softmax-attention layers with linear-attention blocks whose
per-token cost is constant. Qwen3.5 interleaves Gated DeltaNet (GDN) blocks with gated softmax attention
at a 3:1 ratio~\cite{qwen2026qwen35,yang2025gdn}; Kimi Linear uses Kimi Delta Attention with the same
pattern~\cite{kimi2025linear}; Nemotron and Falcon-H1 use Mamba-2 state space blocks
instead~\cite{dao2024mamba2}. Throughout this paper we write $L_a$ for the number of softmax-attention
layers and $L_g$ for the number of linear-attention layers. For Qwen3.5-4B, $L_a=8$ and $L_g=24$ out of
32 layers; for Qwen3.5-0.8B, $L_a=6$ and $L_g=18$ out of 24.

\subsection{Notation}
\label{sec:notation}

Table~\ref{tab:notation} in Appendix~\ref{app:details} lists the symbols used throughout.

\subsection{Two kinds of inference memory}
\label{sec:memory}

During autoregressive decoding a hybrid model maintains two memories with different growth laws.

\paragraph{KV cache.} Each attention layer stores the keys and values of every past token. With $\Hkv$
key-value heads of dimension $d$ and context length $T$, the number of cached elements is
\begin{equation}
\KV(T) = 2 L_a \Hkv d\, T.
\label{eq:kv}
\end{equation}

\begin{figure}[t]
\centering
\includegraphics[width=\linewidth]{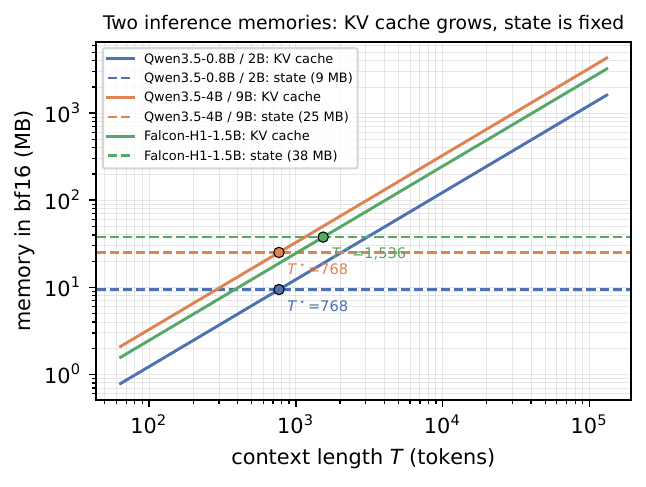}
\caption{KV-cache and recurrent-state size against context length (MB, bf16). $T^\star=768$ for all
Qwen3.5 sizes and 1,536 for Falcon-H1-1.5B.}
\label{fig:crossover}
\end{figure}

\paragraph{Recurrent state.} Each GDN layer compresses the whole past into a fixed-size associative memory
$S\in\mathbb{R}^{H\times d_k\times d_v}$, independent of $T$:
\begin{equation}
|S| = L_g H d_k d_v.
\label{eq:state}
\end{equation}
For Qwen3.5-4B ($\Hkv=4$, $d=256$, $H=32$, $d_k=d_v=128$) this gives $\KV(T)=16{,}384\,T$ and
$|S|\approx 1.26\times10^{7}$ elements, so the two memories are equal at $T^\star\approx768$ tokens.
Below $T^\star$ the recurrent state is the larger memory; above it the KV cache dominates and grows
without bound. Figure~\ref{fig:crossover} plots both quantities. The crossover is the reason a single
quantization policy cannot be optimal across context lengths: the memory worth compressing changes with
$T$.

\subsection{Gated delta rule}
\label{sec:deltarule}

A GDN layer updates its state once per token. Writing $q_t,k_t\in\mathbb{R}^{d_k}$ and
$v_t\in\mathbb{R}^{d_v}$ for the (L2-normalised) query, key and value of token $t$, $\alpha_t\in(0,1)$
for the learned decay gate and $\beta_t\in(0,1)$ for the write strength, the update
is~\cite{schlag2021fwp,yang2025gdn}
\begin{align}
S_t &= \alpha_t S_{t-1} + \beta_t k_t \bigl(v_t - S_{t-1}^{\top} k_t\bigr)^{\top}, \label{eq:gdn}\\
o_t &= S_t^{\top} q_t. \label{eq:gdnout}
\end{align}
The decay $\alpha_t$ forgets the whole state uniformly, while the delta term overwrites only the component
along $k_t$. Both gates are input dependent and are produced by small projections of the hidden state.
In the inference engines we study, $S_t$ is recomputed with a chunked kernel during prefill and with a
per-token recurrent kernel during decoding~\cite{yang2024fla}; in both cases the engine stores $S_t$ in
full precision between steps.

\paragraph{How each memory enters the output.} An attention layer reads its cache only through the
attention weights and the weighted sum,
\begin{equation}
o_t = \mathrm{softmax}\Bigl(\tfrac{1}{\sqrt{d}}\, q_t^{\top} [\,k_1 \cdots k_t\,]\Bigr)\,
[\,v_1 \cdots v_t\,]^{\top},
\label{eq:attn}
\end{equation}
so a rounding error on $k_i$ or $v_i$ perturbs every later output by a term weighted by the attention
that token receives, but the stored value itself is never updated again. A GDN layer, by contrast, reads
and rewrites its state at every step through~\eqref{eq:gdn}, which is what lets errors compound.

\paragraph{Mamba-2 control.} Falcon-H1 replaces the delta rule with a Mamba-2 state-space
block~\cite{dao2024mamba2} whose per-head state follows
\begin{equation}
S_t = \alpha_t S_{t-1} + k_t v_t^{\top},
\label{eq:mamba}
\end{equation}
i.e.\ the same decay but a plain additive write with no erasure along $k_t$. Comparing~\eqref{eq:gdn}
and~\eqref{eq:mamba} contrasts the two update rules: in Mamba-2 the transition is simply
$M_j=\alpha_j I$, with no erasure and overwrite along $k_t$. Section~\ref{sec:mamba} reports the result
of this contrast.

\subsection{Post-training quantization of inference memory}
\label{sec:ptq}

Post-training quantization (PTQ) replaces a stored tensor $x$ by a low-bit code and reconstructs an
approximation $\tilde x$ on read. For the KV cache, per-token or per-channel group quantization at 4 bits
is close to lossless on most benchmarks, and 2-bit schemes recover much of the gap with outlier
handling~\cite{hooper2024kvquant,liu2024kivi}. For recurrent states, the first PTQ studies appeared only
recently and report that even 8-bit storage can hurt and that 4-bit storage is often catastrophic
without compensation~\cite{zhang2026damp,pan2026leapquant,yao2026stepquant}. These works quantize one
memory at a time; the interaction between the two is the subject of this paper.

%% file: sections/04_method.tex
\section{Joint Quantization of KV Cache and Recurrent State}
\label{sec:method}

\subsection{Store-time fake quantization}
\label{sec:storetime}

We study store-time quantization: a tensor is quantized at the moment it is written to the cache and the
dequantized value is what every later step reads. This matches how a serving engine would use low-bit
storage and is implemented, without touching the model, by intercepting the two cache write paths of the
inference library: the KV append of each attention layer and the state write of each GDN layer. Because
our aim is accuracy rather than speed, quantization is simulated (``fake'') in PyTorch; weights,
activations and all arithmetic stay in bf16.

We use asymmetric uniform quantization with a shared scale and zero point per group of $g$ consecutive
elements. For a group $x\in\mathbb{R}^{g}$ and bit-width $b$,
\begin{equation}
\begin{gathered}
Q_b(x) = \Bigl(\mathrm{clamp}\bigl(\lfloor x/s \rceil + z,\, 0,\, 2^{b}-1\bigr) - z\Bigr)\, s,\\
s = \frac{\max x - \min x}{2^{b}-1}, \qquad z = \lfloor -\min x / s \rceil.
\end{gathered}
\label{eq:quant}
\end{equation}
Keys and values are grouped along the head dimension of each token (per-token groups, $g=128$),
following KIVI~\cite{liu2024kivi}. The recurrent state is grouped along $d_v$, so that each row
$S[h,i,:]$ of the associative memory shares one scale. We write $\bkv$ for the KV bit-width (applied to
both K and V) and $\bs$ for the state bit-width; $b=16$ denotes no quantization.

\begin{algorithm}[t]
\caption{One decoding step with store-time quantization}
\label{alg:step}
\KwIn{token $x_t$; stored caches $\tilde K_{<t},\tilde V_{<t}$ (attention layers), $\tilde S_{t-1}$ and
residual $r_{t-1}$ (GDN layers); bit-widths $\bkv,\bs$; flag EF}
\For{each layer $\ell$ in order}{
  \eIf{$\ell$ is an attention layer}{
    compute $k_t,v_t$ from the hidden state\;
    $\tilde k_t \gets Q_{\bkv}(k_t)$, $\tilde v_t \gets Q_{\bkv}(v_t)$ \tcp*{per-token groups}
    append $(\tilde k_t,\tilde v_t)$ to the cache; attend with~\eqref{eq:attn} over
    $\tilde K_{\le t},\tilde V_{\le t}$\;
  }{
    compute $q_t,k_t,v_t,\alpha_t,\beta_t$; $\hat S_t \gets$~\eqref{eq:gdn} applied to $\tilde S_{t-1}$\;
    \lIf{EF}{$\hat S_t \gets \hat S_t + r_{t-1}$}
    $\tilde S_t \gets Q_{\bs}(\hat S_t)$ \tcp*{groups along $d_v$}
    \lIf{EF}{$r_t \gets \hat S_t - \tilde S_t$}
    output $o_t \gets \tilde S_t^{\top} q_t$\;
  }
}
\end{algorithm}

Algorithm~\ref{alg:step} summarises one decoding step with both hooks.

\paragraph{Semantics across prefill and decode.} During prefill the chunked kernel computes the state in
full precision and only the final state is quantized once. During decoding, the state is quantized after
every token. Consequently state-quantization error only accumulates in the decode phase, which is why
all our language-modelling metrics are measured with teacher-forced, token-by-token decoding
(Section~\ref{sec:setup}) rather than a single forward pass.

\subsection{Error propagation in the two memories}
\label{sec:errprop}

The two memories differ in how a rounding error, once committed, affects the future. Let
$\varepsilon^K_t = Q_{\bkv}(k_t) - k_t$ be the error on a cached key. It is bounded by the quantization
step, $\|\varepsilon^K_t\|_\infty \le s_t/2$, it is committed once, and it never changes: KV errors are
independent across tokens and enter the output only through the attention weights of later queries.

For the recurrent state, let $S^\star_t$ be the exact state and $\tilde S_t = Q_{\bs}(\hat S_t)$ the
stored state, where $\hat S_t$ is the result of applying~\eqref{eq:gdn} to the previously stored state
$\tilde S_{t-1}$. Writing $\delta_t = \tilde S_t - \hat S_t$ for the fresh rounding error at step $t$ and
\begin{equation}
M_j = \alpha_j\bigl(I - \beta_j k_j k_j^{\top}\bigr)
\label{eq:transition}
\end{equation}
for the (linear) state transition of the delta rule, the accumulated error obeys
\begin{equation}
\tilde S_t - S^\star_t = \sum_{i\le t} \Bigl(\prod_{j=i+1}^{t} M_j\Bigr)\delta_i.
\label{eq:accum}
\end{equation}
Equation~\eqref{eq:accum} has two consequences. First, each $\delta_i$ is damped by the product of decays
$\prod_j \alpha_j$ and partially erased along the key directions written afterwards, so an individual
error has a finite lifetime whose scale is set by the layer's decay distribution: for a layer with
typical decay $\abar$ an error loses half its norm after $\nhalf = \ln 0.5/\ln\abar$ tokens. Second,
because a new $\delta_t$ is injected at every step, the total error does not decay to zero but converges
to a steady state whose magnitude grows with the quantization step and with $\nhalf$. Layers with slow
decay (large $\abar$) would therefore be expected to accumulate more error for the same bit-width, which
motivates the layer-wise analysis of Section~\ref{sec:layer-sensitivity}.

The steady state can be bounded. Let $\abar_\ell<1$ be an upper bound on the decay of layer $\ell$ and
$\dbar_\ell$ a bound on the per-step rounding error norm (for $b$-bit uniform quantization,
$\dbar_\ell \le \frac{\sqrt{|S_\ell|}}{2}\, s_{\max}$ with $s_{\max}$ the largest group step).

\begin{proposition}[Steady-state error of the quantized delta rule]
\label{prop:steady}
For a GDN layer whose transition matrices satisfy $\|M_j\|_2\le\abar$ for all $j$, with $\abar<1$, and
whose rounding errors satisfy $\|\delta_i\|_F\le\dbar$, the stored state obeys
\begin{equation}
\bigl\|\tilde S_t - S^\star_t\bigr\|_F \le \dbar \sum_{i\le t} \abar^{\,t-i} \le \frac{\dbar}{1-\abar}
\approx \frac{\nhalf\,\dbar}{\ln 2}
\label{eq:bound}
\end{equation}
for $\abar\to1$.
\end{proposition}

The proof (Appendix~\ref{app:proof}) applies the triangle inequality to~\eqref{eq:accum} and uses
$\|I-\beta_j k_j k_j^{\top}\|_2\le1$ for $\beta_j\in[0,1]$ and unit-norm $k_j$. Two readings matter for
allocation. The bound is linear in the quantization step, so lowering $\bs$ by one bit doubles the
ceiling, exactly as for the KV cache; but it is also proportional to the error half-life $\nhalf$, which
is a property of the layer and can span orders of magnitude across layers. The state therefore needs
more bits than the KV cache not because its elements are harder to quantize but because a slowly
decaying layer integrates its rounding errors over $\nhalf$ tokens. Section~\ref{sec:layer-sensitivity}
tests the natural per-layer prediction of this bound, that sensitivity scales with $\nhalf$, and finds
that it does not hold; Section~\ref{sec:additivity} shows why the bound is loose in practice.

The two error sources also interact. The attention layers read hidden states that were produced by GDN
layers carrying $\tilde S_t - S^\star_t$, and the GDN layers read hidden states produced by attention
over quantized keys. If the two errors were independent perturbations of the output distribution, the
loss increase under joint quantization would equal the sum of the increases under each alone. We test
this additivity directly (Section~\ref{sec:additivity}) and, to separate structural interaction from the
mere presence of noise, we include a control in which one memory receives Gaussian noise matched in RMS
to its quantization error instead of the quantization itself.

\subsection{Error feedback for the recurrent state}
\label{sec:ef}

Round-to-nearest storage of a slowly changing state has a known failure mode: when the per-token update
$S_t - S_{t-1}$ is smaller than the quantization step, the stored state freezes while the model keeps
proposing changes, an effect reported for GRU states in low-precision temporal
inference~\cite{erbas2026writeback}. The standard remedy is error
feedback~\cite{karimireddy2019ef,seide2014onebit}: the rounding residual is kept and added back before the
next write,
\begin{equation}
\tilde S_t = Q_{\bs}\bigl(\hat S_t + r_{t-1}\bigr), \qquad r_t = \bigl(\hat S_t + r_{t-1}\bigr) - \tilde S_t,
\label{eq:ef}
\end{equation}
with $r$ reset to zero after prefill. Error feedback costs one full-precision residual buffer per layer
during generation, i.e.\ $16|S|$ bits per active request on top of the $\bs|S|$ bits of the quantized
state. For a serving engine this overhead is transient (it lives with the request, not in the paged
cache) but it is not free, and we report it alongside every EF result. Every configuration in this paper
is evaluated both with plain rounding (RTN) and with error feedback (EF), because the answer to ``where
should the bits go'' changes depending on whether the state is protected.

\subsection{Metrics}
\label{sec:metrics}

\paragraph{Decode-time loss.} For a document $x_{1:T}$ we prefill the first $P$ tokens in one pass, then
feed the remaining tokens one at a time under the quantized caches and record the teacher-forced negative
log-likelihood
\begin{equation}
\mathrm{NLL} = -\frac{1}{T-P}\sum_{t=P+1}^{T} \log p\bigl(x_t \mid \tilde x_{<t}\bigr),
\label{eq:nll}
\end{equation}
where the conditioning is on the stored caches. We report
$\dnll = \mathrm{NLL}(\bkv,\bs) - \mathrm{NLL}(16,16)$ in nats, averaged over documents, with paired
bootstrap confidence intervals over documents (the same documents are used for every configuration).

\paragraph{Additivity index.} For a joint configuration we define
\begin{equation}
\mathcal{I}(\bkv,\bs) = \dnll(\bkv,\bs) - \dnll(\bkv,16) - \dnll(16,\bs).
\label{eq:additivity}
\end{equation}
$\mathcal{I}=0$ means the two errors act independently on the loss; $\mathcal{I}>0$ is super-additive
interaction. We compute $\mathcal{I}$ per document and test whether its confidence interval excludes
zero.

\paragraph{Memory and Pareto frontier.} The footprint of a configuration at context length $T$ is
$\bkv\KV(T) + \bs|S|$ bits (plus the EF residual when used). A configuration is Pareto-optimal if no
other configuration has both lower footprint and lower loss. Loss is measured at $P+D=512$ tokens; we
report the frontier at $T\in\{512, 4\mathrm{k}, 32\mathrm{k}\}$, where the 4k and 32k footprints are
extrapolated with~\eqref{eq:kv} and the loss is held at its 512-token value. The frontier at long context
therefore shows how the memory axis moves, not how the loss changes with $T$.

\paragraph{Task accuracy.} For reasoning we report exact-match accuracy on GSM8K in thinking mode together
with mean generated tokens, the fraction of generations that hit the token budget, and an 8-gram
repetition rate as a loop indicator. For retrieval we report needle-in-a-haystack accuracy by context
length and needle depth.

\subsection{The bit-allocation problem}
\label{sec:allocation}

Given a total memory budget $B$ (in bits) for the two caches at context length $T$, the allocation problem
is
\begin{equation}
\min_{\bkv,\bs}\ \mathcal{L}(\bkv,\bs) \quad \text{s.t.} \quad \bkv\KV(T) + \bs|S| \le B,
\label{eq:alloc}
\end{equation}
where $\mathcal{L}$ is the task loss (negative log-likelihood, or one minus accuracy). Prior work fixes
one of the two bit-widths at 16 and optimises the other, so its feasible set is a one-dimensional slice
of ours; the joint optimum can therefore never be worse. The empirical question is how much better it
is, and whether the optimum moves with $T$ (through~\eqref{eq:kv}), with model scale, with the task, and
with the presence of error feedback. We answer it by measuring $\mathcal{L}$ on a grid of $(\bkv,\bs)$
pairs and reading off the Pareto frontier of memory against loss, rather than by
solving~\eqref{eq:alloc} analytically.

%% file: sections/05_setup.tex
\section{Experimental Setup}
\label{sec:setup}

\paragraph{Models.} Qwen3.5 at 0.8B, 2B, 4B and 9B parameters. The 0.8B and 2B models share one cache
geometry ($L=24$, $L_a=6$, $L_g=18$, $\Hkv=2$, $d=256$, $H=16$, $d_k=d_v=128$: $\KV/\text{token}=6{,}144$,
$|S|=4.72$M, $T^\star=768$), and so do 4B and 9B ($L=32$, $L_a=8$, $L_g=24$, $\Hkv=4$, $H=32$:
$\KV/\text{token}=16{,}384$, $|S|=12.58$M, $T^\star=768$). Falcon-H1-1.5B is the Mamba-2 control.
Appendix~\ref{app:details} gives the full table.

\paragraph{Hardware and software.} All runs use one NVIDIA A100-SXM4-40GB in bf16 with PyTorch 2.11 and
\texttt{transformers} 5.19.0.dev0 (commit \texttt{66e0f4a}). The GDN recurrence runs in the library's
pure-PyTorch kernels; no \texttt{fla} or \texttt{causal-conv1d} kernels are used, so the per-token state
write is the same code path in every configuration. Both hooks quantize with group size 128; the EF
residual is reset after prefill. Quantization is calibration-free: scale and zero point are computed at
store time from each group's range, and no data is used to fit them. For Falcon-H1 the library updates
the Mamba-2 state in place during single-token decoding, so we re-store it through the cache after every
token to apply the same store-time quantization as for GDN; we also correct a batched-decoding bug in the
library's gated RMSNorm. Both corrections are in the released code (\texttt{hybridquant/compat.py}).

\paragraph{Language modelling.} WikiText-2 test (and C4 validation for the second-corpus check) cut into
non-overlapping windows of 512 tokens, $P=256$ prefill and $D=256$ teacher-forced decode tokens. A
longer-decoding check uses $P=256$ and $D=1{,}024$ on 3 seeds $\times$ 10 different windows of 1,280
tokens (Section~\ref{sec:longdecode}). Each configuration is run on 3 seeds $\times$ 20 windows (the
single-layer sweep of Section~\ref{sec:layer-sensitivity} uses 1 seed $\times$ 8 windows); $\dnll$ is
paired per document against the bf16 run of the same window, and confidence intervals are paired
bootstraps over documents with 1,000 resamples.

\paragraph{Reasoning.} GSM8K test, a fixed 200-problem subset drawn with seed 0 (indices released with
the code), thinking mode, 4,096-token generation budget, sampling parameters from the model card
(temperature 1.0, top-$p$ 0.95, top-$k$ 20, min-$p$ 0). Accuracy intervals are bootstrap 95\% intervals
over problems. Differences between configurations are tested with an exact McNemar test on the paired
problems.

\paragraph{Retrieval.} Single-needle haystacks from WikiText-2 training text at 4k and 16k tokens, five
depths $\times$ ten needles per length ($n=50$), greedy decoding, 32 generated tokens; a response is
correct if it contains the six-character code.

\paragraph{Grid.} $\bkv \in \{16,8,4,3,2\} \times \bs \in \{16,8,6,5,4\} \times \{\mathrm{RTN},\mathrm{EF}\}$
on 0.8B; the core cells (4-bit and 8-bit of each memory, the 4/4 corner and the recipe cells) on 2B, 4B
and 9B.

%% file: sections/06_results.tex
\section{Results}
\label{sec:results}

\subsection{The grid: a large and stable asymmetry}
\label{sec:grid}

\begin{figure*}[t]
\centering
\includegraphics[width=0.49\textwidth]{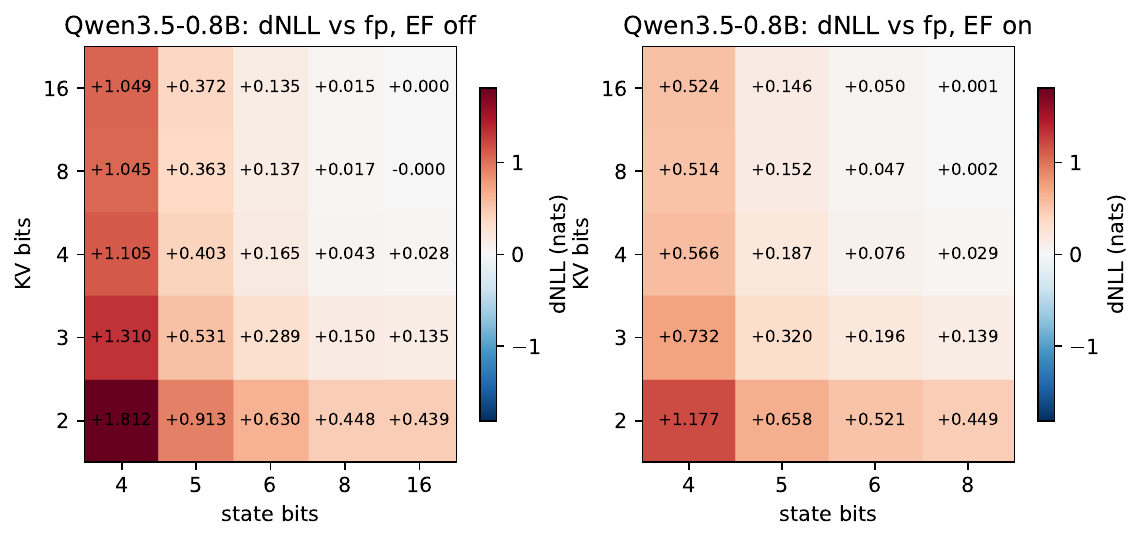}
\includegraphics[width=0.49\textwidth]{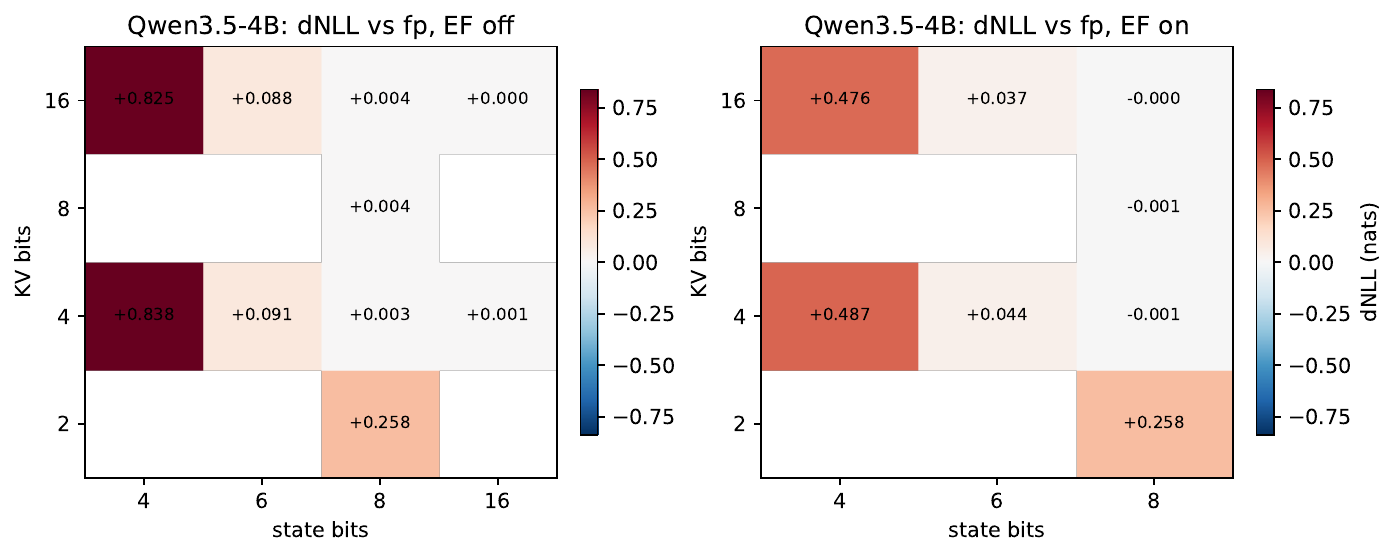}
\caption{$\dnll$ (nats) over the $(\bkv,\bs)$ grid, RTN (left of each pair) and error feedback (right),
for Qwen3.5-0.8B (full grid) and 4B (core cells; blank = not run). 3 seeds $\times$ 20 windows.}
\label{fig:heatmap}
\end{figure*}

\begin{figure*}[t]
\centering
\includegraphics[width=\textwidth]{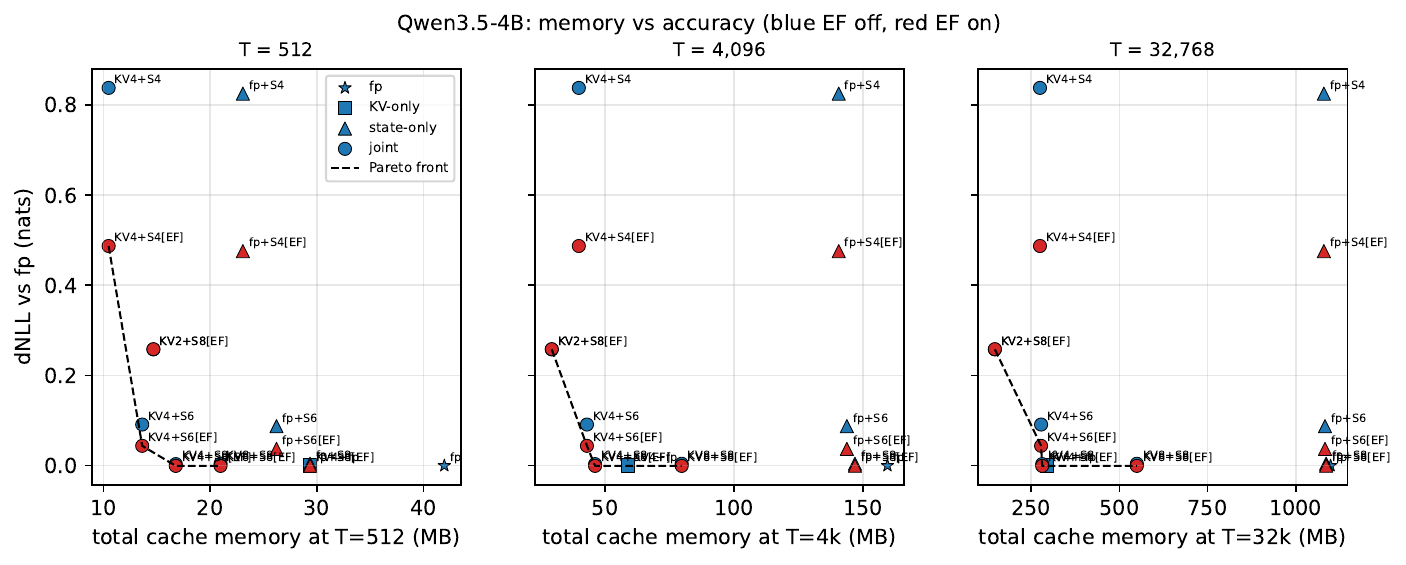}
\caption{Cache footprint against $\dnll$ on Qwen3.5-4B at $T=512$, 4k and 32k. Loss is measured at 512
tokens; footprint from eqs.~\ref{eq:kv}--\ref{eq:state}, without the EF residual. Shape: setting;
colour: EF off (blue) or on (red); dashed: Pareto front. KV4+S8 (16.8~MB) lies on the front with no
measurable loss.}
\label{fig:pareto}
\end{figure*}

Figure~\ref{fig:heatmap} shows the full grid on 0.8B and the core cells on 4B. The dominant feature is
the asymmetry between the two memories. On 0.8B, 4-bit KV costs $+0.028$ nats $[0.024, 0.032]$ over a
bf16 NLL of 2.974, while 4-bit state costs $+1.049$ $[1.001, 1.094]$, thirty-seven times more for the
same bit-width. The 8-bit state costs $+0.015$ $[0.010, 0.020]$. At 4B the pattern is the same and
sharper: 4-bit KV is indistinguishable from bf16 ($+0.001$ $[-0.003, 0.004]$), 4-bit state costs
$+0.825$ $[0.751, 0.902]$, and 8-bit state costs $+0.004$ $[0.001, 0.007]$, statistically
distinguishable from zero but about four times below the 8-bit cost on 0.8B. Reading down the state axis
on 0.8B with round-to-nearest, the state costs $+0.372$ $[0.347, 0.395]$ at 5 bits and $+0.135$
$[0.122, 0.148]$ at 6 bits; with error feedback on, it is unusable at 4 bits ($+0.524$), marginal at
5 bits ($+0.146$), usable at 6 bits ($+0.050$) and lossless at 8 bits ($+0.001$). Along the KV axis,
8-bit storage is exactly lossless ($-0.000$) and the loss stays small until 2 bits, where it jumps to
$+0.439$ $[0.416, 0.464]$ on 0.8B; on 4B, 2-bit KV with an 8-bit state costs $+0.258$
$[0.239, 0.279]$. The 8-bit state is thus the first bit-width that is lossless, and the 4-bit KV cache is
lossless or nearly so at every size we tested.

\subsection{Error feedback}
\label{sec:ef-results}

Error feedback is the largest lever that leaves the stored format and the paged-cache footprint
unchanged. On 0.8B it halves the 4-bit state loss, from $+1.049$ to $+0.524$ (KV in bf16), cuts the
5-bit loss from $+0.372$ to $+0.146$ and the 6-bit loss from $+0.135$ to $+0.050$, and lowers the joint
$(\mathrm{KV4},\mathrm{S4})$ cell from $+1.105$ to $+0.566$. Averaged over all cells of the grid in which the state is
quantized, EF lowers $\dnll$ by 0.222 nats on 0.8B (20 cells), 0.440 on 2B (6 cells), 0.101 on 4B
(8 cells, range $-0.351$ to $+0.000$) and 0.045 on 9B (6 cells); the absolute benefit shrinks with scale
because the RTN loss it repairs shrinks with scale (Section~\ref{sec:scale}). At 8 bits EF is
unnecessary on NLL (0.8B: $+0.001$ with EF versus $+0.015$ without). The cost is the $16|S|$-bit residual
per active request, which is why the recipe in Section~\ref{sec:recipe} reports EF cells with and without
that transient buffer.

\subsection{Memory against accuracy: the recipe}
\label{sec:recipe}

Figure~\ref{fig:pareto} turns the 4B grid into a frontier. At $T=512$ the bf16 caches occupy 41.9~MB. The
joint cell $(\mathrm{KV4},\mathrm{S6})$ with error feedback costs $+0.044$ $[0.036, 0.053]$ nats at 13.6~MB, 32\% of bf16;
without EF the same cell costs $+0.091$ $[0.081, 0.101]$. On GSM8K the same recipe cell keeps accuracy at
0.555 against 0.600 for bf16; it loses 35 problems and gains 26, a difference that is not significant
(McNemar $p=0.31$). The $(\mathrm{KV4},\mathrm{S8})$ cell sits at 40\% (16.8~MB by~\eqref{eq:kv} and~\eqref{eq:state}); on
Qwen3.5-4B it is lossless, $+0.003$ $[-0.001, 0.007]$ nats with RTN and $-0.001$ $[-0.004, 0.003]$ with
EF, with needle-in-a-haystack accuracy 1.00 at 4k and 16k; $(\mathrm{KV8},\mathrm{S8})$ at 21.0~MB (50\%) costs $+0.004$
$[0.001, 0.007]$ with RTN, like the 8-bit state alone, and $-0.001$ $[-0.002, 0.001]$ with error
feedback. On 0.8B the same cell costs $+0.043$ nats with RTN and $+0.029$ with EF. By comparison, the
best one-memory configurations, KV-only at 4 bits and state-only at 8 bits, each stop at 70\% of the bf16
footprint, because at $T=512$ the untouched memory is at least as large as the one being compressed. The
joint optimum is therefore not marginally better than the one-dimensional slices that prior work
explores; on Qwen3.5-4B it reaches 40\% of bf16 with no measurable loss where either slice stops at
70\%. Table~\ref{tab:headline} collects the headline cells. At $T=4$k and 32k the KV cache dominates the
footprint (Figure~\ref{fig:crossover}) and the state bits barely move the memory axis: at 32k with the KV
cache in bf16, the difference between a 6-bit and a 16-bit state is 1.4\% of the total, and with the KV
cache at 4 bits it is about 5\%, so the state should be the last memory to lose bits at long context, and
the KV cache the first.

\subsection{Scale}
\label{sec:scale}

The asymmetry holds at every scale (Appendix Figure~\ref{fig:scale}). The 4-bit state penalty is
$+1.049$ (0.8B), $+1.808$ (2B), $+0.825$ (4B) and $+0.283$ (9B); the 4-bit KV penalty is $+0.028$,
$+0.011$, $+0.001$ (n.s.) and $+0.006$; the 8-bit state penalty is $+0.015$, $+0.010$, $+0.004$ and
$+0.003$ (n.s.). The state penalty falls with scale, as one would expect if larger models have more
redundant state capacity, but 2B is an outlier with a penalty nearly twice that of 0.8B. This is not a
cache-size effect: 0.8B and 2B have exactly the same $\KV$ per token and the same $|S|$
(Section~\ref{sec:setup}), so the 2B model stores the same number of elements at the same precision.
What differs is the hidden size, which is twice as large at 2B, so the 2B model packs a wider residual
stream through a state of the same capacity. The excess is confined to 4-bit round-to-nearest: with error
feedback the 4-bit state costs $+0.524$ on 0.8B and $+0.577$ on 2B, nearly equal, and at 6 bits RTN 2B is
below 0.8B ($+0.090$ versus $+0.135$). This points to the freezing effect of Section~\ref{sec:ef}, where
updates below the 4-bit step are lost, being stronger at 2B; the mechanism is left open
(Section~\ref{sec:discussion}).

\subsection{Second corpus}
\label{sec:c4}

The 0.8B core cells on C4 validation reproduce WikiText-2 cell for cell: over a bf16 NLL of 3.154, 4-bit
state costs $+1.031$ $[0.975, 1.086]$, 4-bit KV $+0.023$ $[0.019, 0.027]$, 8-bit state $+0.025$
$[0.021, 0.030]$, and the mean EF effect is again $-0.222$ nats (6 cells). The C4 core-cell table is in
Appendix~\ref{app:results}.

\subsection{Task dependence}
\label{sec:task}

\begin{figure*}[t]
\centering
\includegraphics[width=0.49\textwidth]{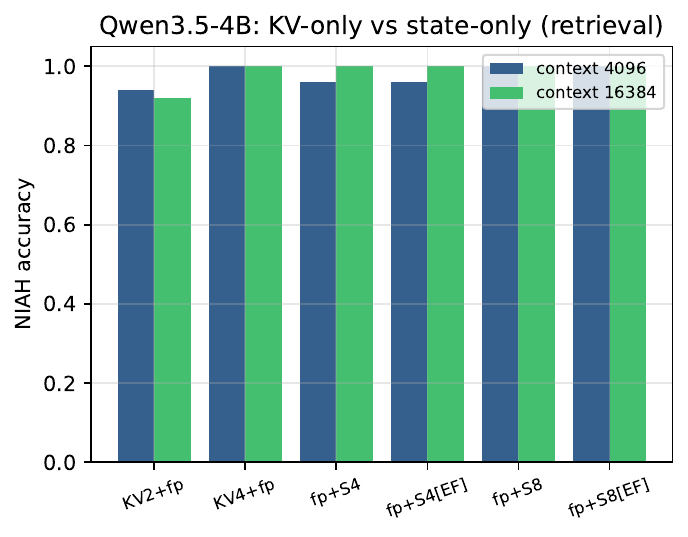}
\includegraphics[width=0.49\textwidth]{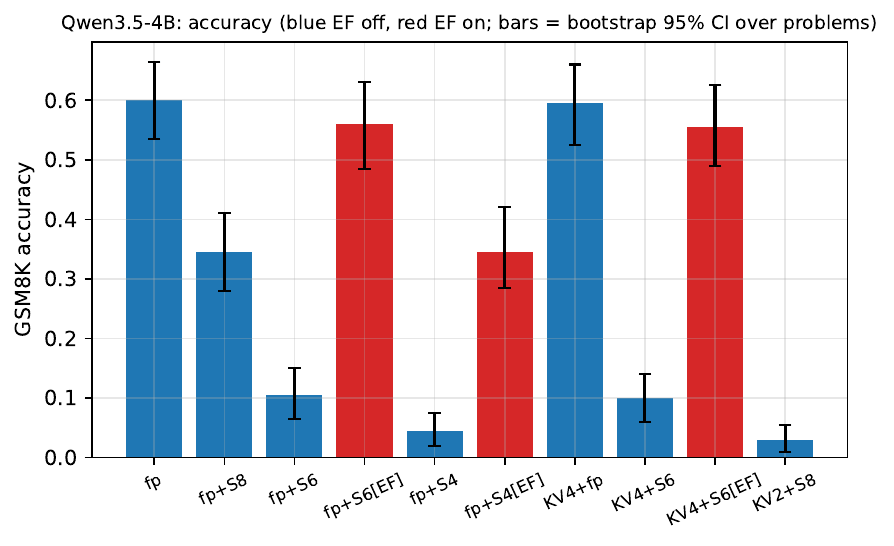}
\caption{Task dependence on Qwen3.5-4B. Left: needle-in-a-haystack accuracy at 4k and 16k ($n=50$ each)
with only the KV cache or only the state quantized. Right: GSM8K accuracy, $n=200$ per bar; red bars use
error feedback; bootstrap 95\% intervals.}
\label{fig:task}
\end{figure*}

The memory that limits accuracy depends on what the model is asked to do (Figure~\ref{fig:task}).
This is consistent with cache ablation and swapping experiments on the same model families, in which
exact retrieval depends on the attention cache~\cite{afendulev2026recalls}.

\paragraph{Retrieval is bounded by the KV cache.} On 4B, needle-in-a-haystack accuracy is 1.00 at both
4k and 16k in bf16, with 4-bit KV, and with 8-bit state. Quantizing the state to 4 bits, the setting that
costs 0.8 nats on NLL, leaves retrieval at 0.96 (4k) and 1.00 (16k), with or without error feedback, and
the joint $(\mathrm{KV4},\mathrm{S4})$ cell at 0.98 and 1.00. The only configuration that hurts is 2-bit KV: 0.94 and 0.92
on 4B, and 0.32 and 0.54 on 0.8B, where bf16 is 1.00 at both lengths. Retrieval reads the needle back
through attention over the cached keys, so it is the fidelity of the KV cache, not of the state, that
decides whether the code survives.

\paragraph{Reasoning is bounded by the state.} On GSM8K in thinking mode, 4B scores 0.600
$[0.535, 0.665]$ in bf16 ($n=200$), generating 2,912 tokens on average, with 48.5\% of generations
hitting the 4,096-token budget. Four-bit KV is lossless (0.595 $[0.525, 0.660]$). The 8-bit state, which
costs $+0.004$ nats on NLL at this scale, lowers accuracy to 0.345 $[0.280, 0.410]$. Because the run uses
one seed at temperature 1.0, we compare problems one by one with an exact McNemar test: the 8-bit state
loses 67 problems that bf16 solves and gains 16 ($p=1.4\times10^{-8}$), whereas 4-bit KV loses 30 and
gains 29 ($p=1.0$). The claim is scoped to bit-widths that are lossless on NLL: at those widths the state
breaks reasoning and the KV cache does not. The KV cache is not irrelevant to reasoning, since 2-bit KV
with an 8-bit state collapses accuracy to 0.030; it is robust down to 4 bits.

The failure mode is visible in the generations (Appendix Figures~\ref{fig:niah-behaviour}
and~\ref{fig:gsm8k-length}). Failing cells do not produce short wrong answers; they run to the token
budget and repeat themselves. With a 6-bit state, 99.5\% of generations hit the 4,096-token budget
against 48.5\% for bf16, and the 8-gram repetition rate is 0.205 to 0.372 for the round-to-nearest cells
that fail, against 0.086. A long chain of thought re-stores the state thousands of times, so a rounding
error that is invisible over a 256-token decode window has thousands of tokens to derail the reasoning
into long, looping generations.

Error feedback repairs most of the damage: it raises the 6-bit state from 0.105 to 0.560, the joint
$(\mathrm{KV4},\mathrm{S6})$ cell from 0.100 to 0.555 and the 4-bit state from 0.045 to 0.345, and brings the
budget-exhaustion and repetition rates of the 6-bit cells back near bf16 (57\% and 0.092 for the 6-bit
state with EF). Neither 6-bit EF cell is distinguishable from bf16 (6-bit state $p=0.37$, joint
$p=0.31$). At 4 bits the repair is partial: error feedback recovers accuracy to 0.345, but the
repetition rate stays at 0.375, so it restores correct answers without removing the looping; at 6 bits
it removes both. A 6-bit state with error feedback (0.560) also beats an 8-bit state without it (0.345):
compared directly, 60 problems are solved only by the former and 17 only by the latter
($p=9\times10^{-7}$). On reasoning, error feedback is worth more than two extra bits. It also restores the
8-bit state exactly to bf16: 0.600, with 32 problems lost and 32 gained ($p=1.0$), and against the 8-bit
state without it, 65 problems solved only with error feedback and 14 only without
($p=5.2\times10^{-9}$). It removes the budget exhaustion and the looping as well (50.0\% and 0.086, equal
to bf16). With error feedback, every state bit-width from 6 bits up is statistically indistinguishable
from bf16 on GSM8K.

\subsection{Longer decoding}
\label{sec:longdecode}

The language-modelling metric decodes 256 tokens, whereas the GSM8K generations re-store the state
thousands of times. To connect the two we repeat the core cells with 1,024 decode tokens on 0.8B and 4B
(Appendix Figure~\ref{fig:longdecode} and tables). These windows differ from those of the 512-token grid,
so we compare the size of the change rather than individual digits. Under round-to-nearest the state
penalty grows with the decode length: on 4B the 4-bit state goes from $+0.825$ to $+1.187$, the 6-bit
state from $+0.088$ to $+0.116$ and the 8-bit state from $+0.004$ to $+0.011$; on 0.8B the 6-bit penalty
nearly doubles ($+0.135$ to $+0.253$) and the 8-bit penalty more than triples ($+0.015$ to $+0.051$), as
expected if rounding error keeps accumulating. With error feedback the 6- and 8-bit penalties do not grow
(4B: $+0.037$ to $+0.038$ and $-0.000$ to $+0.000$; 0.8B: $+0.050$ to $+0.053$ and $+0.001$ to
$+0.004$), while the 4-bit penalty grows only modestly ($+0.476$ to $+0.562$ on 4B, $+0.524$ to $+0.613$
on 0.8B). The KV penalty does not change with length ($+0.001$ and $+0.001$ on 4B, $+0.028$ and $+0.027$
on 0.8B), and the joint $(\mathrm{KV4},\mathrm{S4})$ excess stays non-significant on 4B ($+0.003$ $[-0.019, 0.025]$). This
links the language-modelling results to the GSM8K failure, and it is the strongest evidence that error
feedback stops the accumulation rather than only lowering its level.

\subsection{A Mamba-2 control}
\label{sec:mamba}

On Falcon-H1-1.5B (3 seeds $\times$ 20 documents, bf16 NLL 2.550; Appendix Figure~\ref{fig:falcon}) the
KV side of the asymmetry is unchanged but the state side disappears. The KV cache behaves as in Qwen3.5:
4-bit KV costs $+0.010$ $[0.004, 0.015]$ nats and 2-bit KV with an 8-bit state costs $+0.205$. The
Mamba-2 state, however, tolerates 4-bit storage with no measurable loss ($+0.004$ $[-0.010, 0.018]$),
where the GDN state loses 0.8 to 1.8 nats, and every state cell from 4 to 8 bits, with and without error
feedback, is within $\pm0.02$ nats of bf16. A few of these cells are slightly below bf16 with intervals
that exclude zero (8-bit $-0.014$, 6-bit $-0.018$); we report them as within $\pm0.02$ nats and do not
interpret them. The joint $(\mathrm{KV4},\mathrm{S4})$ excess is $-0.002$ $[-0.008, 0.005]$, and error feedback has no
effect (mean change $+0.012$ nats over 6 cells). This is consistent with the delta rule, which erases
and overwrites the state along $k_t$ in~\eqref{eq:gdn}, being what makes the GDN state fragile, as
Section~\ref{sec:background} anticipated. Falcon-H1 differs from Qwen3.5 in more than the update rule,
however: every layer is a parallel attention and Mamba-2 block, its state dimension is larger (256), and
the model and its training recipe differ. The comparison therefore supports, but does not isolate, the
delta rule as the cause.

%% file: sections/07_analysis.tex
\section{Analysis}
\label{sec:analysis}

\subsection{Additivity and the noise-matched control}
\label{sec:additivity}

\begin{figure*}[t]
\centering
\includegraphics[width=0.49\textwidth]{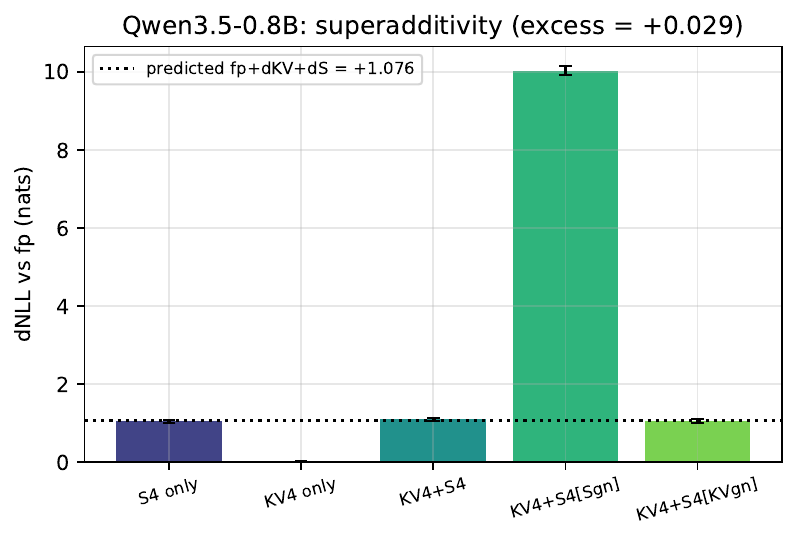}
\includegraphics[width=0.49\textwidth]{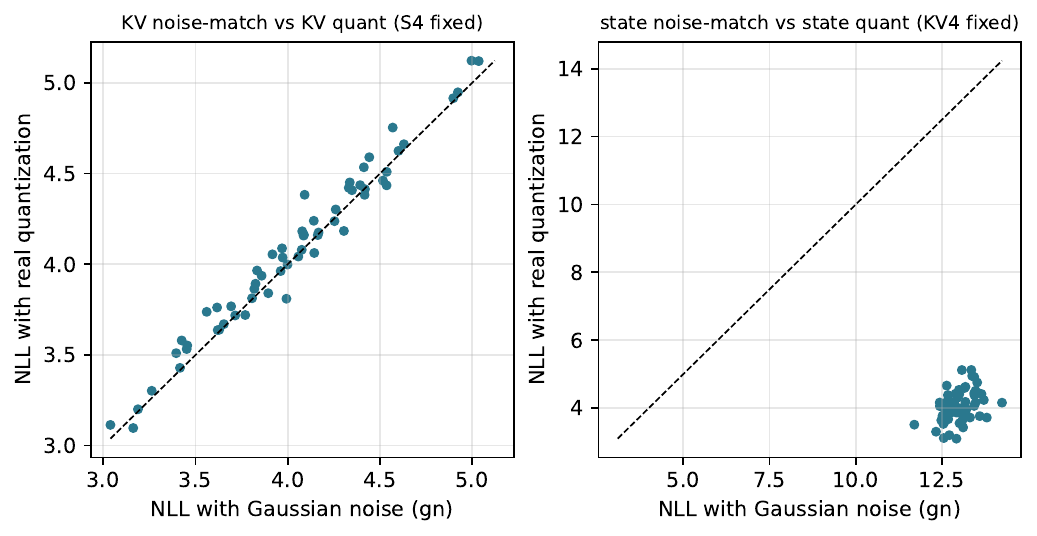}
\caption{Additivity on Qwen3.5-0.8B (3 seeds $\times$ 20 documents). Left: 4-bit state-only, KV-only and
joint cells, then the noise-matched controls (Sgn, KVgn); dotted line: additive prediction; paired
bootstrap 95\% CIs. Right: per-document NLL under quantization against matched Gaussian noise, for the
KV cache and the state; diagonal = equality.}
\label{fig:additivity}
\end{figure*}

\begin{figure*}[t]
\includegraphics[width=0.245\textwidth]{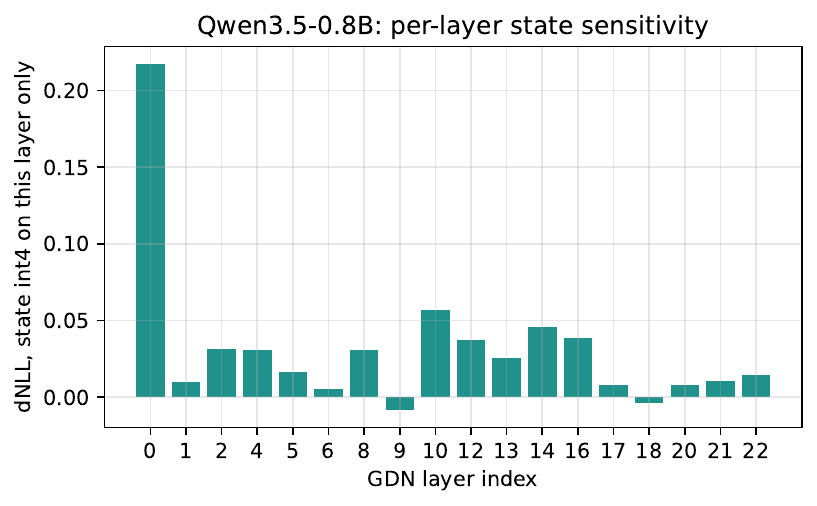}\hfill
\includegraphics[width=0.245\textwidth]{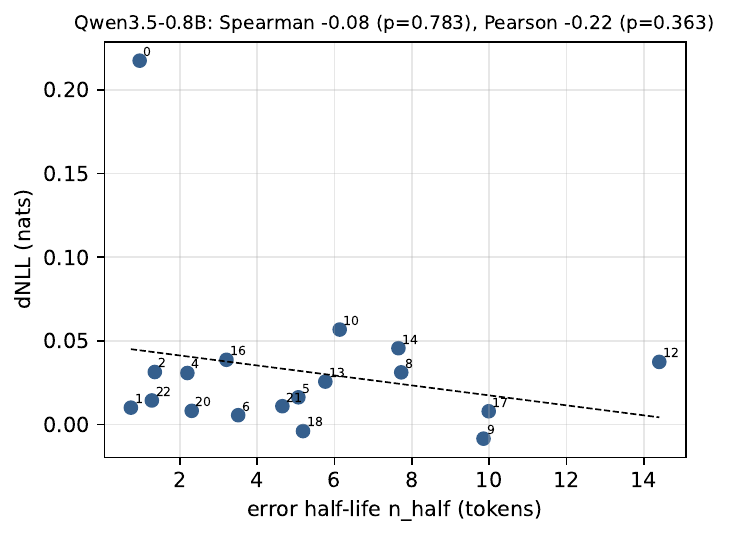}\hfill
\includegraphics[width=0.245\textwidth]{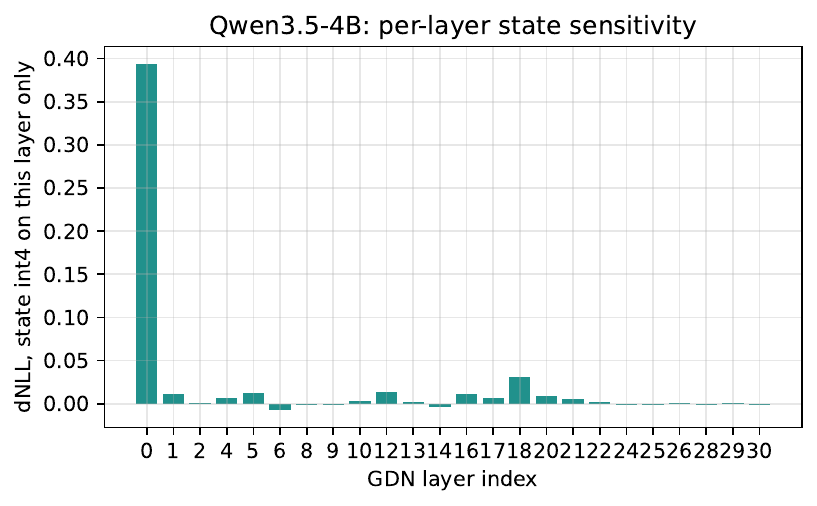}\hfill
\includegraphics[width=0.245\textwidth]{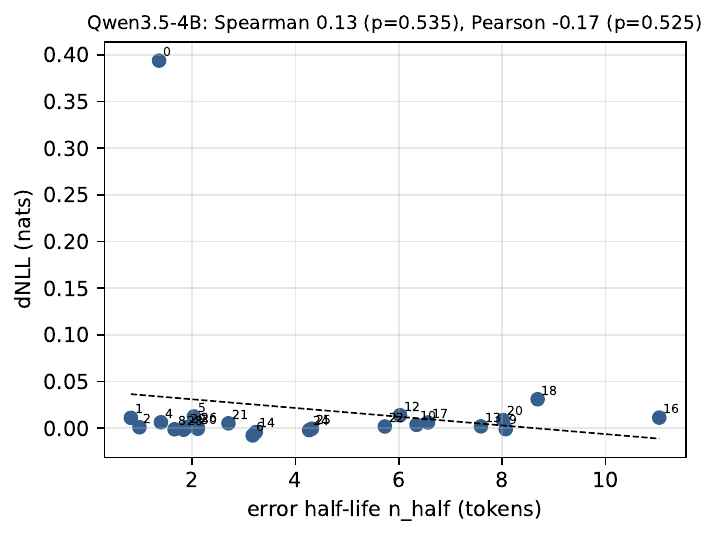}
\caption{Per-layer sensitivity on Qwen3.5-0.8B (left pair) and 4B (right pair), one GDN layer's state at
4 bits (RTN), 1 seed $\times$ 8 documents, no error bars. Second panel of each pair: against the error
half-life $\nhalf$; Spearman $-0.08$ ($p=0.78$) and $+0.13$ ($p=0.54$).}
\label{fig:layers}
\end{figure*}

Figure~\ref{fig:additivity} (left, first three bars) tests whether the two errors add. At the $(\mathrm{KV4},\mathrm{S4})$
corner on 0.8B the joint loss exceeds the sum of the single-memory losses by $+0.029$ nats
$[0.009, 0.049]$; the excess is $+0.047$ $[0.021, 0.073]$ on 2B, $+0.040$ $[0.019, 0.060]$ on 0.8B with
C4, and $-0.002$ $[-0.009, 0.005]$ on 9B; on 4B the index at $(\mathrm{KV4},\mathrm{S6})$ is $+0.003$ (no paired interval was
computed for this cell). The interaction is therefore real but small, about 3\% of the joint loss at
small scale, and it vanishes at 9B. Relative to the one-nat state penalty it is negligible for
allocation: the loss surface over $(\bkv,\bs)$ is, to within a few hundredths of a nat, the sum of two
one-dimensional curves, and the two memories can be allocated almost independently.

The noise-matched control (left, last two bars; right, per document) separates the structure of the
error from its energy. Replacing the 4-bit KV quantization by Gaussian noise of equal RMS, with the state
held at 4 bits, costs $+1.061$ nats against $+1.105$ for quantization: for the KV cache only the error
energy matters, as expected for errors that are committed once and read through a softmax. Replacing the
4-bit state quantization by equal-RMS noise, with the KV cache held at 4 bits, costs $+10.03$ nats
against the same $+1.105$: nine times more. The difference is the idempotence of rounding. Once a state
has been rounded to the 4-bit grid, re-applying~\eqref{eq:gdn} with a small update and rounding again
often returns the same stored value, so no new error is injected at that step; independent noise, by
contrast, is injected at every token and accumulates through~\eqref{eq:accum} exactly as
Proposition~\ref{prop:steady} assumes. The bound of~\eqref{eq:bound} is therefore an accurate description
of the noise control and a loose one of actual quantization, which explains both why the state tolerates
rounding far better than its worst case and why the per-layer prediction of the bound fails (next
section). It also explains the failure mode that error feedback repairs: the same idempotence that
suppresses error injection also freezes a slowly moving state.

\subsection{Layer sensitivity}
\label{sec:layer-sensitivity}

Proposition~\ref{prop:steady} predicts that layers with longer error half-lives should be more sensitive
to state quantization. Figure~\ref{fig:layers} tests this on 0.8B and 4B by storing the state of one GDN
layer at a time at 4 bits (1 seed $\times$ 8 documents, so individual bars carry no interval and only the
ordering and the size of the layer-0 effect should be read). Two things stand out. First, sensitivity is
extremely concentrated: layer 0 costs $+0.2175$ nats on its own, 38\% of the summed single-layer effects
($+0.575$), while every other layer is below $+0.06$ and the least sensitive (layer 9) is $-0.0085$,
i.e.\ zero within noise. Second, the sum of the single-layer effects ($+0.575$) is about half of the
all-layer effect ($+1.049$), so the layers interact super-additively with each other even though the two
memories interact only weakly. Against the prediction, the correlation between $\nhalf$ and sensitivity
is absent: Spearman $-0.08$ (permutation $p=0.78$), Pearson $-0.22$ ($p=0.36$). The measured half-lives
span roughly 1 to 14 tokens, and layer 0, the most sensitive layer, has one of the shortest. The decay
half-life does not order the layers. The finding replicates and is stronger on 4B (1 seed $\times$ 8
documents): layer 0 costs $+0.394$ nats alone, 80\% of the summed single-layer effects ($+0.495$), the
next largest is layer 18 at $+0.031$, layer 0 again has one of the shortest half-lives (1.36 tokens), and
the correlation with $\nhalf$ is again absent (Spearman $+0.13$, permutation $p=0.54$). The dominance of
the first GDN layer, whose state is written from raw token embeddings and read by every layer above it,
points instead to position in the network as the relevant variable, and to layer 0 as the obvious
candidate for a higher bit-width in a mixed-precision scheme. Jiang~\cite{jiang2026scale} reports the
same concentration on Qwen3.5-0.8B and 2B (63 to 78\% of the summed single-layer damage at 4 bits) and
finds that simply protecting the first three layers matches a learned allocation; our sweep adds the 4B
scale and the null result against the half-life prediction. We leave a mixed-precision scheme to future
work.

\subsection{State format ablation}
\label{sec:ablation}

Appendix Figure~\ref{fig:ablation} and the ablation table in Appendix~\ref{app:results} vary the state
quantizer on 0.8B with the KV cache fixed at 4 bits (RTN, 3 seeds $\times$ 20 documents). Group size is
the strongest lever: shrinking the group along $d_v$ from 128 to 64 to 32 lowers the 4-bit state cost
from $+1.105$ to $+0.519$ to $+0.304$ nats, and the 6-bit cost from $+0.165$ to $+0.096$ to $+0.062$. A
group of 32 is a $3.6\times$ reduction at 4 bits, larger than error feedback buys, at a price of roughly
one extra bit per element in scale and zero-point overhead. Symmetric scales are three times worse than
asymmetric ones at 4 bits ($+3.341$), as expected for a state whose rows are not centred. Stochastic
rounding is the worst option of all, five times the deterministic cost ($+5.248$ at 4 bits, $+0.851$ at
6 bits). This is the mechanism of Section~\ref{sec:additivity} observed a second way: stochastic rounding
injects a fresh random error at every token, which is exactly what the noise-matched control does and
exactly what round-to-nearest, being idempotent, avoids. Deterministic rounding with error feedback is
the right combination for a recurrent state; stochastic rounding, which is often recommended for weights
and gradients, should not be used for it. A smaller group and error feedback are complementary, since
one changes the stored format and the other the write rule; their combination was not run.

%% file: sections/08_discussion.tex
\section{Discussion and Limitations}
\label{sec:discussion}

\paragraph{Where the bits should go.} The results support a simple rule for GDN hybrids that is consistent
across language modelling, retrieval and reasoning. First, keep error feedback on the state at all times:
its residual lives with the request rather than in the paged cache, so it costs no cache memory, and
without it even an 8-bit state lowers GSM8K accuracy from 0.600 to 0.345 ($p=1.4\times10^{-8}$). Second,
the default setting is 4-bit KV with a 6-bit state and error feedback: on Qwen3.5-4B it uses 32\% of the
bf16 cache footprint at 512 tokens, costs $+0.044$ nats, and keeps GSM8K at 0.555, not distinguishable
from bf16 ($p=0.31$); its retrieval accuracy is 1.00. Third, for a lossless setting use 4-bit KV with an
8-bit state and error feedback at 40\%: on Qwen3.5-4B it costs $-0.001$ nats with retrieval at 1.00, and
its components are lossless on every task measured for them (the 8-bit state with error feedback scores
0.600 on GSM8K, $p=1.0$, and 4-bit KV 0.595, $p=1.0$); the joint cell itself was not evaluated on GSM8K.
An 8-bit state without error feedback should not be treated as safe for reasoning. If memory must go
lower, reduce the state bits under error feedback rather than taking the KV cache to 2 bits ($+0.44$
nats on 0.8B, a measurable retrieval loss and GSM8K 0.030). At long context the KV cache is the footprint
and the state is a rounding error on the memory axis, so bits should be removed from the KV cache first
and the state should keep its bits.

\paragraph{Limitations.} (i) Quantization is simulated; we measure accuracy, not speed, and a real low-bit
state kernel has to reproduce the store-time semantics of Algorithm~\ref{alg:step}, including the
residual buffer for error feedback. (ii) The language-modelling metric uses 256 decode tokens, checked at
1,024 on two sizes; longer horizons remain untested apart from the GSM8K generations, which are a single
task on one model size, with one seed at temperature 1.0 on a 200-problem subset. With one seed at
temperature 1.0, configurations whose net difference to bf16 is below roughly 15 problems (0.075) are not
distinguishable (McNemar $p>0.05$ for KV4, S6+EF and KV4+S6+EF). (iii) The study covers one GDN family
at four sizes and one Mamba-2 control at one size. (iv) The 2B anomaly in Section~\ref{sec:scale} is
unexplained, but confined to 4-bit RTN. (v) Only uniform asymmetric group quantization is studied; the
intra-memory schemes of DAMP, STEPQuant and KIVI are orthogonal and could be composed with our split.

%% file: sections/09_conclusion.tex
\section{Conclusion}
\label{sec:conclusion}

In Gated DeltaNet hybrids the recurrent state is the fragile memory and the KV cache the robust one:
4-bit KV costs almost nothing at every scale from 0.8B to 9B, 4-bit state costs about a nat, and the two
errors add almost exactly, so the budget can be split per memory rather than searched jointly. The state
is the fragile memory for reasoning as well, and error feedback is what makes a low-bit state usable
there: without it even an 8-bit state lowers GSM8K accuracy from 0.600 to 0.345. The resulting recipe
keeps error feedback on and uses 4-bit KV with a 6-bit state, which on Qwen3.5-4B reaches 32\% of the
bf16 cache for 0.04 nats and keeps GSM8K at 0.555, not distinguishable from bf16; 4-bit KV with an 8-bit
state is lossless at 40\% on Qwen3.5-4B. In a Mamba-2 hybrid the same 4-bit state costs nothing, which
points to the delta rule as the source of the fragility. Two findings remain open: a single early GDN
layer carries most of the state sensitivity at both 0.8B and 4B, uncorrelated with its decay half-life,
and the 2B model is 1.7 times as sensitive as 0.8B and 2.2 times as sensitive as 4B at 4-bit RTN despite
sharing its cache geometry with 0.8B, an effect that is unexplained but confined to 4-bit RTN.

\section*{Acknowledgments}

This work was carried out by a single author without institutional funding; compute was self-funded on
Google Colab. The author used Claude (Anthropic) as an assistant for code development and manuscript
editing; all experiments, analyses and claims were designed, run and verified by the author.

\section*{Impact Statement}

This work aims to reduce the memory cost of serving language models, which lowers the hardware needed
for a given workload. It does not introduce new capabilities and we foresee no specific negative impact
beyond those of more efficient LLM inference in general.

\section*{Reproducibility and Data Availability}

All configurations, raw results and the scripts that regenerate every table and figure from them are
available at \url{https://github.com/skytinstone/hybridquant}. Each table is produced by
\texttt{python scripts/run\_grid.py -{}-config configs/<name>.yaml}, each report by
\texttt{python analysis/report.py -{}-stage <name> -{}-inputs results/<name>.jsonl}, and the whole
pipeline by \texttt{bash scripts/run\_all.sh}. All datasets used are public (WikiText-2, C4, GSM8K). The
author declares no conflict of interest and no external funding.

%% file: sections/appendix.tex
\section{Proof of Proposition~\ref{prop:steady}}
\label{app:proof}

\noindent From~\eqref{eq:accum} and the triangle inequality,
$\|\tilde S_t - S^\star_t\|_F \le \sum_{i\le t} \big\|\prod_{j=i+1}^{t} M_j\big\|_2 \|\delta_i\|_F$.
Each factor satisfies $\|M_j\|_2 = \alpha_j \|I-\beta_j k_j k_j^{\top}\|_2$. For unit-norm $k_j$ the
matrix $k_j k_j^{\top}$ is a rank-one projector with eigenvalues $\{1,0,\dots\}$, so
$I-\beta_j k_j k_j^{\top}$ has eigenvalues $\{1-\beta_j,1,\dots\}$ and spectral norm
$\max(1,|1-\beta_j|)=1$ for $\beta_j\in[0,1]$. Hence $\|M_j\|_2\le\alpha_j\le\abar$ and
$\big\|\prod_{j=i+1}^{t} M_j\big\|_2 \le \abar^{\,t-i}$. Summing the geometric series
gives~\eqref{eq:bound}; the final approximation uses $1-\abar\approx-\ln\abar=\ln2/\nhalf$ for $\abar$
close to 1.\qed

The bound is loose in three ways that matter in practice: it ignores the erasure of error components
along later keys (which can only help); it treats all heads of a layer with one $\abar$ whereas the decay
is input-dependent and head-specific; and, as Section~\ref{sec:additivity} shows empirically, it assumes
a fresh error of norm $\dbar$ at every token, whereas round-to-nearest is idempotent and injects no new
error when the update does not cross a rounding boundary. The noise-matched control is the setting in
which the third assumption holds, and there the measured loss is an order of magnitude larger than under
rounding.

\section{Experimental details}
\label{app:details}

\begin{table}[h]
\centering
\caption{Notation.}
\label{tab:notation}
\small
\begin{tabular}{ll}
\toprule
Symbol & Meaning \\
\midrule
$L_a$, $L_g$ & number of attention / GDN layers \\
$\Hkv$, $d$ & KV heads and head dimension of attention \\
$H$, $d_k$, $d_v$ & GDN heads, key and value dimensions \\
$T$ & context length (prefill + generated tokens) \\
$T^\star$ & crossover length where $\KV=|S|$ \\
$S_t\in\mathbb{R}^{H\times d_k\times d_v}$ & recurrent state after token $t$ \\
$\alpha_t$, $\beta_t$ & decay gate and write strength of the delta rule \\
$Q_b(\cdot)$, $g$ & $b$-bit group quantizer, group size \\
$\bkv$, $\bs$ & bit-width of KV cache / recurrent state \\
$\tilde x$, $x^\star$ & stored (quantized) and exact value of $x$ \\
$\delta_t$, $r_t$ & fresh rounding error, error-feedback residual \\
$M_j$ & state transition of the delta rule, Eq.~\eqref{eq:transition} \\
$\nhalf$ & error half-life in tokens, $\ln0.5/\ln\abar$ \\
$\Delta\mathcal{L}$ & loss increase relative to bf16 \\
$B$ & total memory budget in bits \\
\bottomrule
\end{tabular}
\end{table}

\paragraph{Models.}
\begin{center}
\resizebox{\textwidth}{!}{\input{tables/models}}
\end{center}
\noindent For Falcon-H1 every layer is a parallel hybrid block that holds both an attention head group and a
Mamba-2 state, so $L_a=L_g=L$; $d_k$ and $d_v$ are the Mamba-2 head and state dimensions.

\paragraph{Language-modelling data.} WikiText-2 test and C4 validation, concatenated and cut into
non-overlapping windows of $P+D$ tokens; $P=256$ prefill and $D=256$ decode tokens unless stated;
20 windows per seed, 3 seeds. The window indices for each seed are fixed by the config and released with
the code.

\paragraph{Reasoning.} GSM8K test, a fixed 200-problem subset sampled with seed 0 (indices released with
the code), thinking mode, prompt ``Please reason step by step, and put your final answer within
{\ttfamily\char92 boxed\char123\char125}.'', generation capped at 4,096 new tokens; sampling parameters from the
model card: temperature 1.0, top-$p$ 0.95, top-$k$ 20, min-$p$ 0, no presence penalty.

\paragraph{Retrieval.} Single-needle haystack built from concatenated WikiText-2 training text; the
needle is a sentence containing a random six-character code absent from the corpus; depths
$\{0.1,0.3,0.5,0.7,0.9\}$, ten needles per depth, contexts of 4k and 16k tokens; greedy decoding,
32 generated tokens; a response is correct if it contains the code.

\paragraph{Hardware and software.} One NVIDIA A100-SXM4-40GB; bf16; PyTorch 2.11; \texttt{transformers}
5.19.0.dev0 (commit \texttt{66e0f4a}); pure-PyTorch GDN kernels, no \texttt{fla} or
\texttt{causal-conv1d}.

\paragraph{Compute.} Language-modelling and retrieval stages of rounds 1 and 2 took about 5.4 A100 GPU
hours, GSM8K 13.2 hours (4.25~h for the first 432 generations at batch 8 and 8.95~h for the remaining
1,568 at batch 16), and round 3 about 4.4 hours, for about 23 A100 GPU hours in all valid runs (per-stage
table in \texttt{paper/tables/compute.md}). Two invalid Falcon-H1 runs (1.6 hours) are excluded.

\section{Full results}
\label{app:results}

Every table in this appendix is generated by \texttt{analysis/make\_tex\_tables.py} from the
\texttt{results/*.jsonl} files; the values are mean $\dnll$ with paired bootstrap 95\% CIs over
documents, RTN and EF side by side.

\noindent
\begin{minipage}[t]{0.542\textwidth}
\textbf{Qwen3.5-0.8B, WikiText-2}\\[2pt]
\resizebox{\linewidth}{!}{\input{tables/grid_0.8b}}
\end{minipage}\hfill
\begin{minipage}[t]{0.431\textwidth}
\textbf{State format ablation (0.8B)}\\[2pt]
\resizebox{\linewidth}{!}{\input{tables/ablation_0.8b}}
\end{minipage}

\vspace{7pt}
\noindent
\begin{minipage}[t]{0.492\textwidth}
\textbf{Qwen3.5-0.8B, C4}\\[2pt]
\resizebox{\linewidth}{!}{\input{tables/grid_c4_0.8b}}
\end{minipage}\hfill
\begin{minipage}[t]{0.492\textwidth}
\textbf{Qwen3.5-2B}\\[2pt]
\resizebox{\linewidth}{!}{\input{tables/grid_2b}}
\end{minipage}

\vspace{7pt}
\noindent
\begin{minipage}[t]{0.492\textwidth}
\textbf{Qwen3.5-4B}\\[2pt]
\resizebox{\linewidth}{!}{\input{tables/grid_4b_core}}
\end{minipage}\hfill
\begin{minipage}[t]{0.492\textwidth}
\textbf{Qwen3.5-9B}\\[2pt]
\resizebox{\linewidth}{!}{\input{tables/grid_9b}}
\end{minipage}

\pagebreak
\paragraph{Falcon-H1-1.5B (Mamba-2).}

\begin{center}
\small
\input{tables/grid_falcon_1.5b}
\end{center}

\noindent
\begin{minipage}[t]{0.492\textwidth}
\textbf{Longer decoding, Qwen3.5-0.8B ($D=1{,}024$)}\\[2pt]
\resizebox{\linewidth}{!}{\input{tables/longdecode_0.8b}}
\end{minipage}\hfill
\begin{minipage}[t]{0.492\textwidth}
\textbf{Longer decoding, Qwen3.5-4B ($D=1{,}024$)}\\[2pt]
\resizebox{\linewidth}{!}{\input{tables/longdecode_4b}}
\end{minipage}

\medskip

\begin{figure}[h]
\centering
\includegraphics[width=0.8\textwidth]{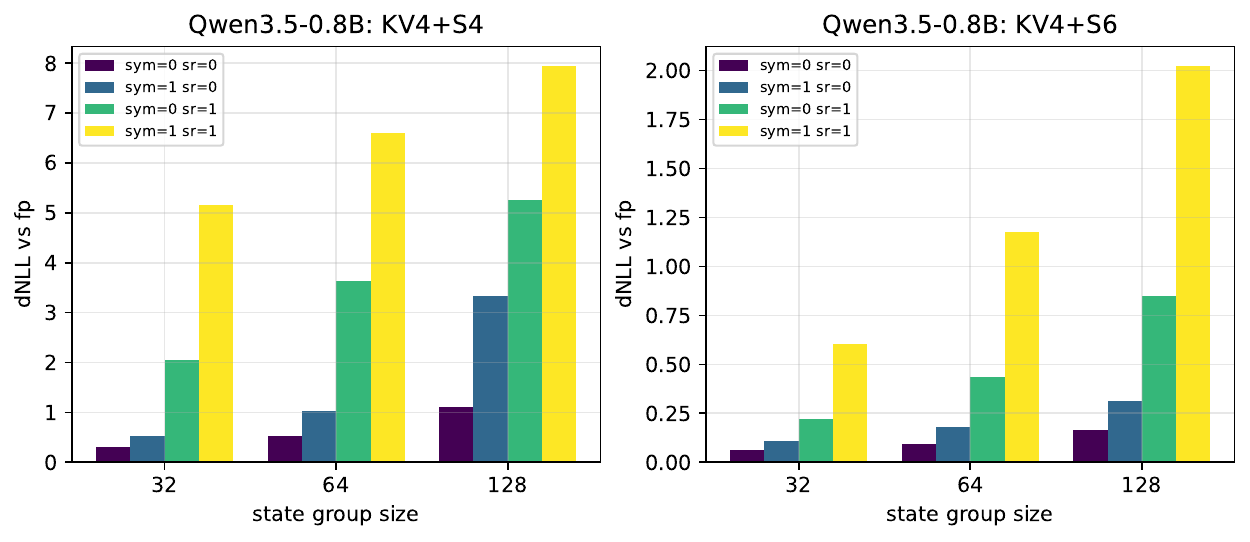}
\caption{State format ablation on Qwen3.5-0.8B, KV at 4 bits, RTN: group size, symmetric scales (sym)
and stochastic rounding (sr), at 4-bit (left) and 6-bit (right) state. 3 seeds $\times$ 20 documents.}
\label{fig:ablation}
\end{figure}

\begin{figure}[h]
\centering
\includegraphics[width=0.44\textwidth]{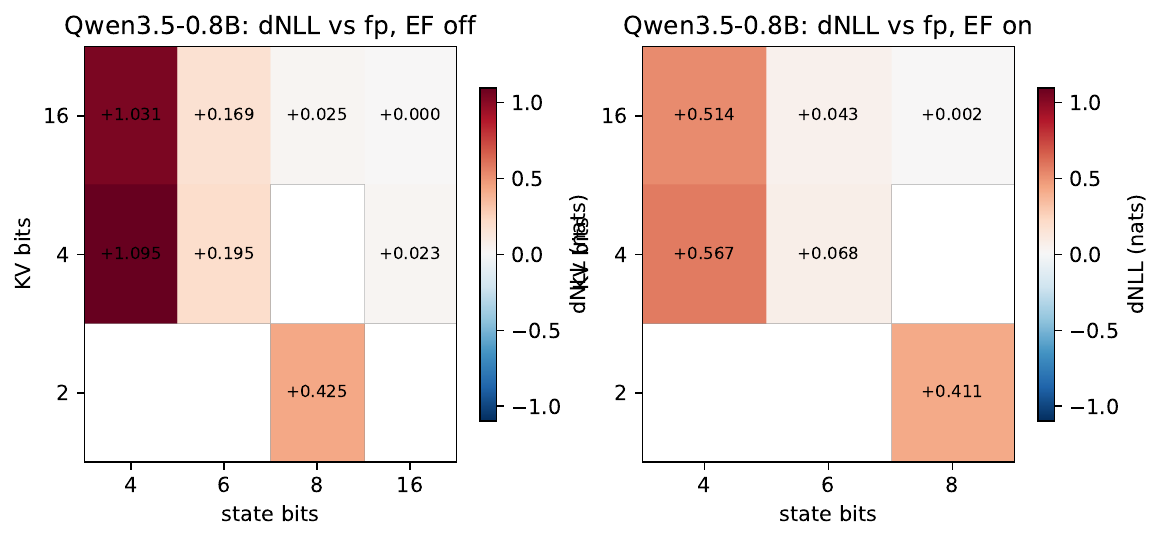}
\includegraphics[width=0.44\textwidth]{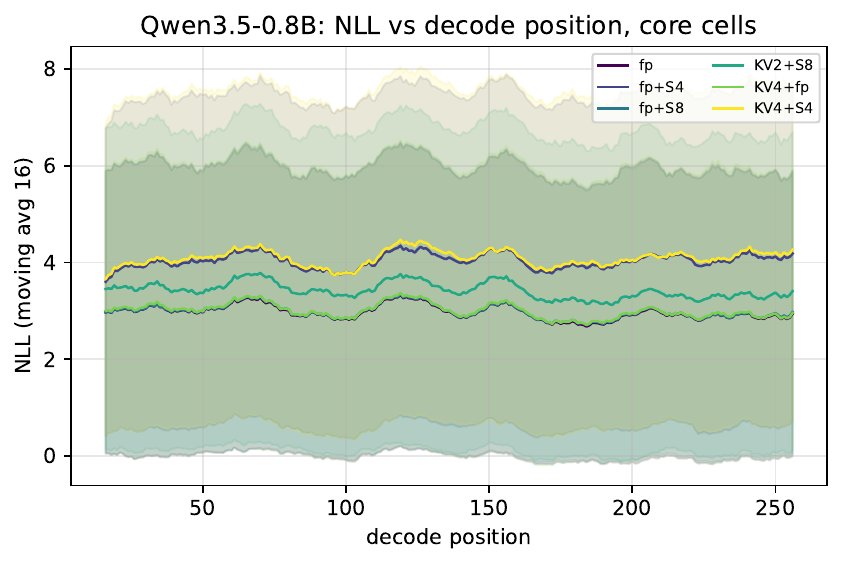}
\caption{Left: 0.8B core cells on C4, RTN and EF. Right: NLL (16-token moving average) against decode
position on 0.8B; shading $\pm1$ SD across documents.}
\label{fig:c4}
\end{figure}

\begin{figure}[p]
\centering
\includegraphics[width=0.44\textwidth]{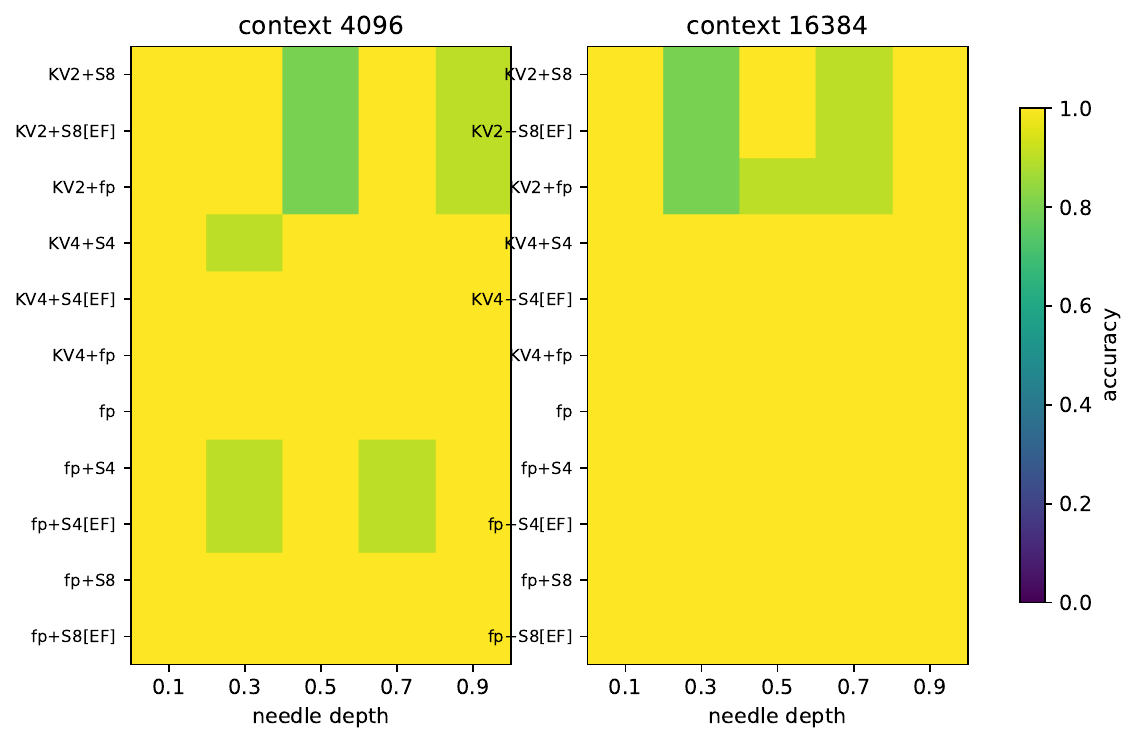}
\includegraphics[width=0.44\textwidth]{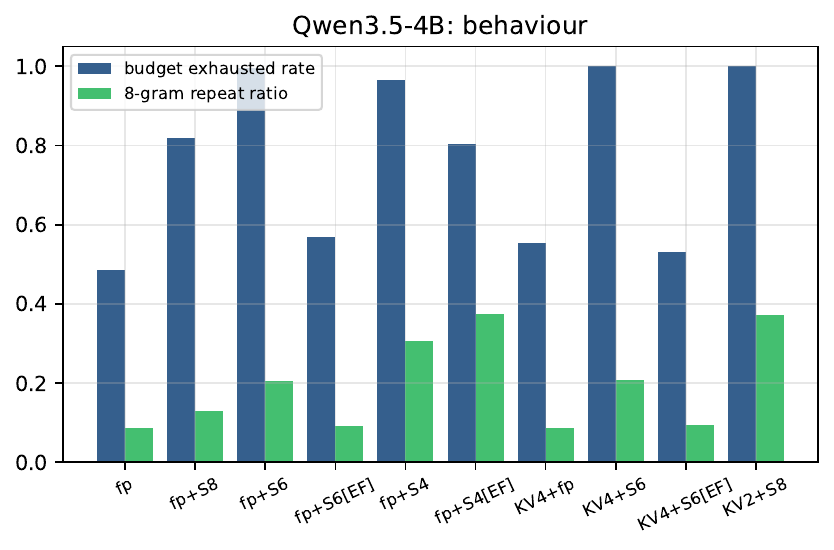}
\caption{Left: NIAH accuracy by needle depth on Qwen3.5-4B (ten needles per cell). Right: GSM8K
budget-exhaustion and 8-gram repetition rates, $n=200$ each.}
\label{fig:niah-behaviour}
\end{figure}

\begin{figure}[p]
\centering
\includegraphics[width=0.44\textwidth]{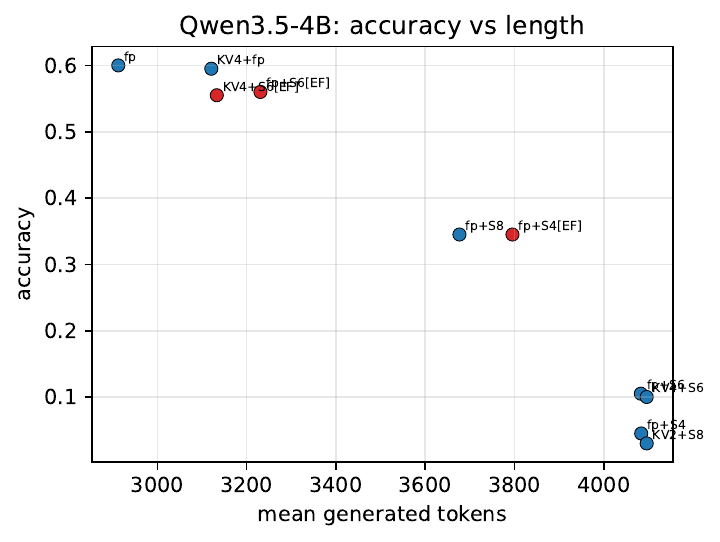}
\includegraphics[width=0.44\textwidth]{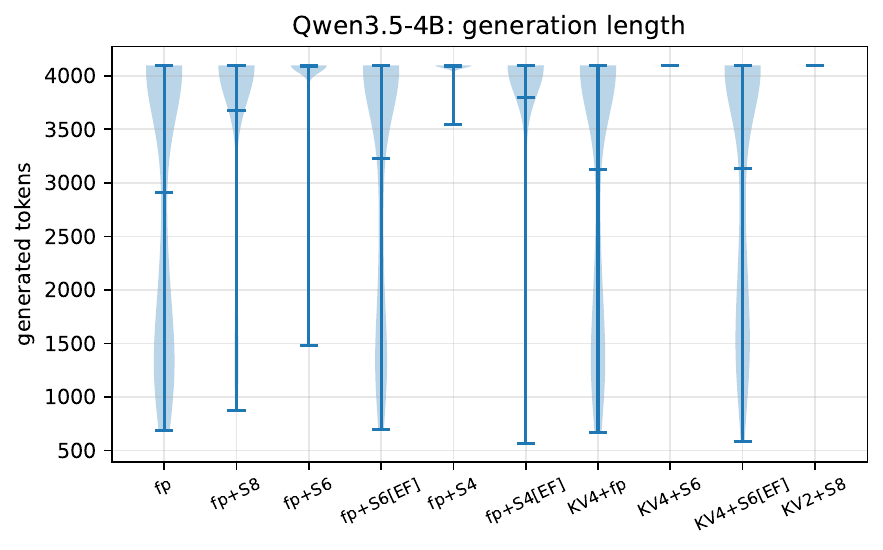}
\caption{GSM8K on Qwen3.5-4B, $n=200$ per configuration. Left: accuracy against mean generated tokens
(red: EF). Right: generated-token distributions; failing configurations cluster at the 4,096-token
budget.}
\label{fig:gsm8k-length}
\end{figure}

\begin{figure}[p]
\centering
\includegraphics[width=0.62\textwidth]{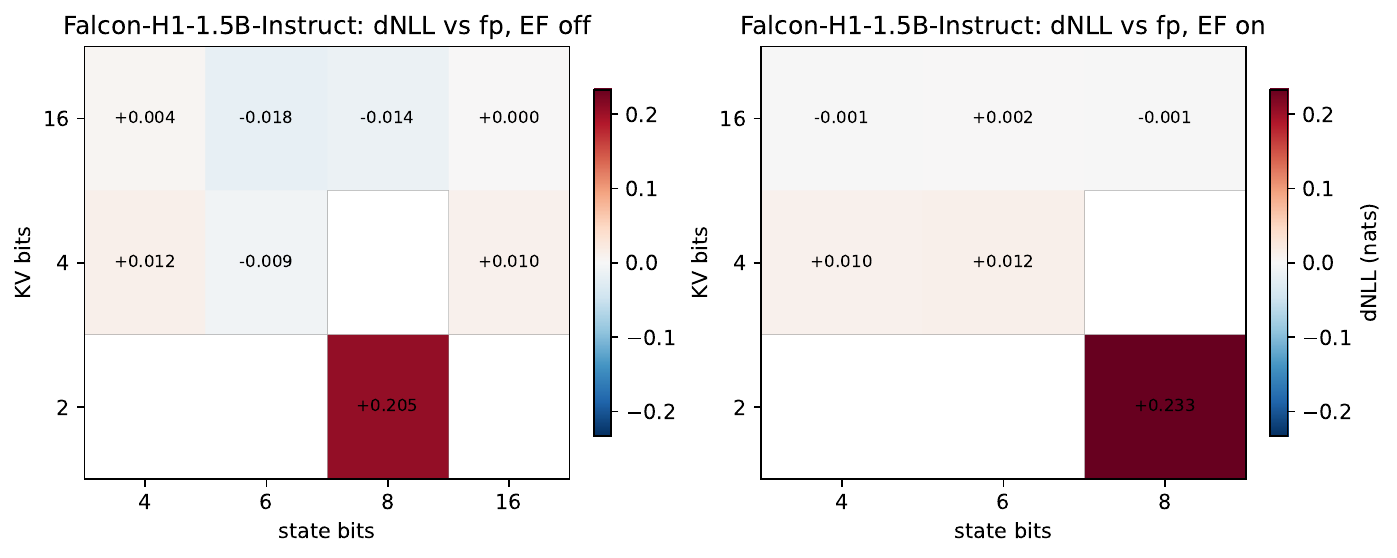}
\includegraphics[width=0.34\textwidth]{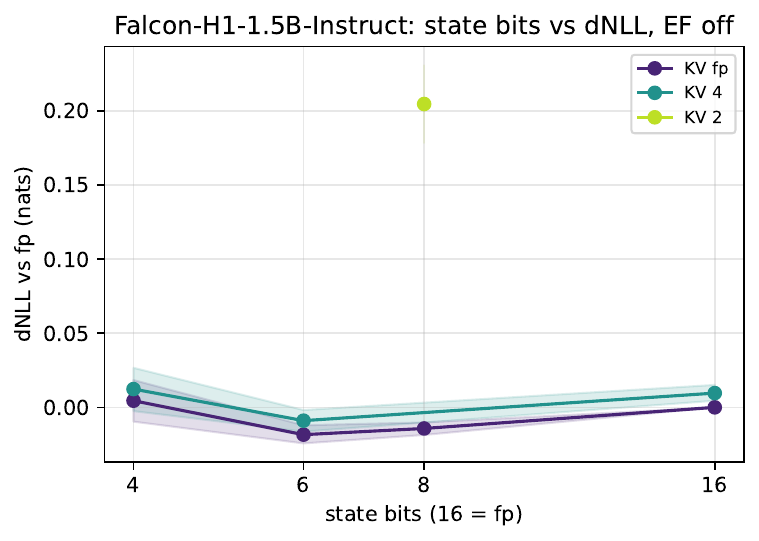}
\caption{Mamba-2 control, Falcon-H1-1.5B, 3 seeds $\times$ 20 documents. Left: $\dnll$ over the core
cells, RTN and EF. Right: $\dnll$ against state bits, RTN, with paired bootstrap 95\% CIs.}
\label{fig:falcon}
\end{figure}

\begin{figure}[p]
\centering
\includegraphics[width=0.44\textwidth]{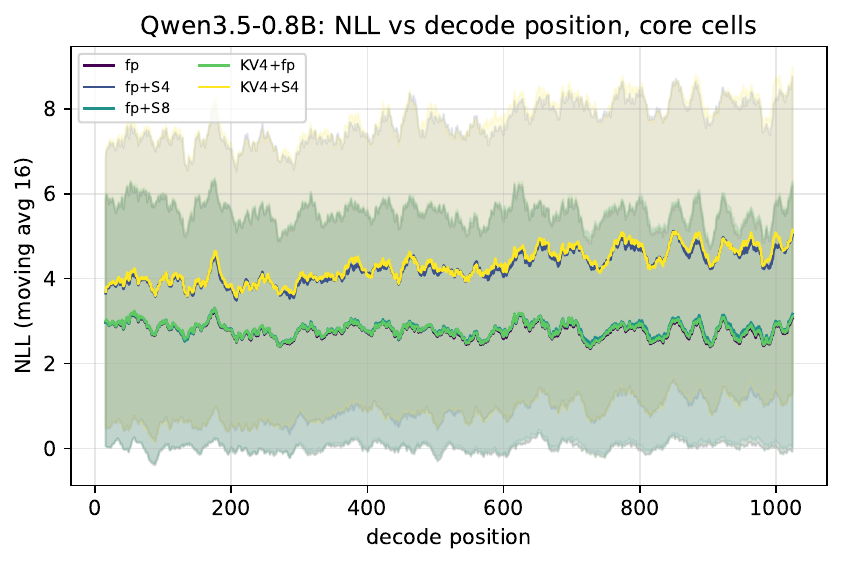}
\includegraphics[width=0.44\textwidth]{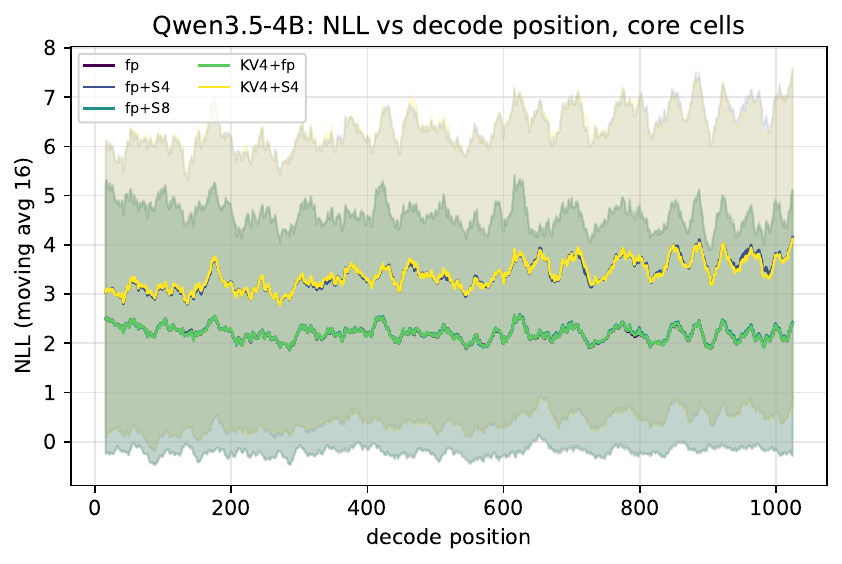}
\caption{NLL (16-token moving average) against decode position with $P=256$, $D=1{,}024$ on Qwen3.5-0.8B
(left) and 4B (right), RTN core cells; 3 seeds $\times$ 10 windows; shading $\pm1$ SD.}
\label{fig:longdecode}
\end{figure}

\begin{figure}[p]
\centering
\includegraphics[width=0.32\textwidth]{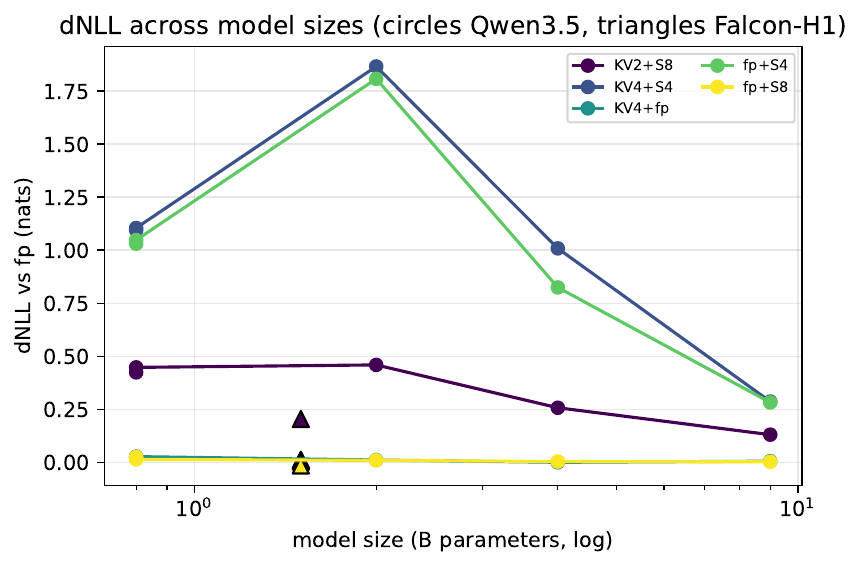}\hfill
\includegraphics[width=0.32\textwidth]{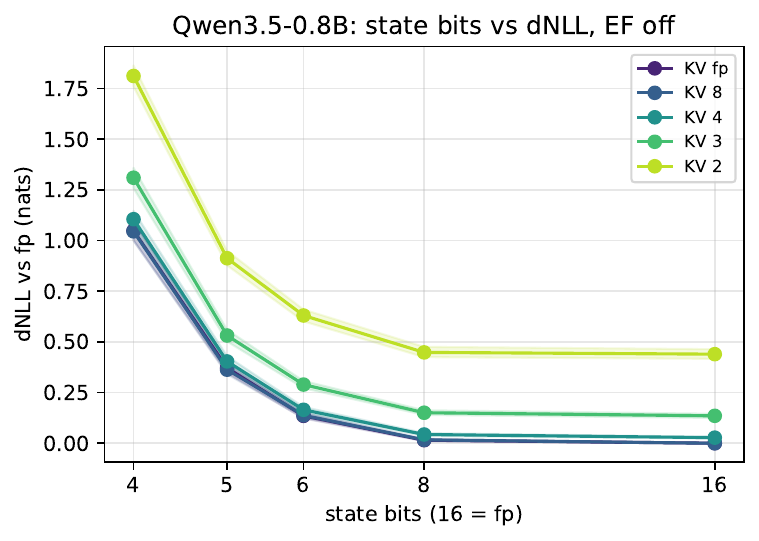}\hfill
\includegraphics[width=0.32\textwidth]{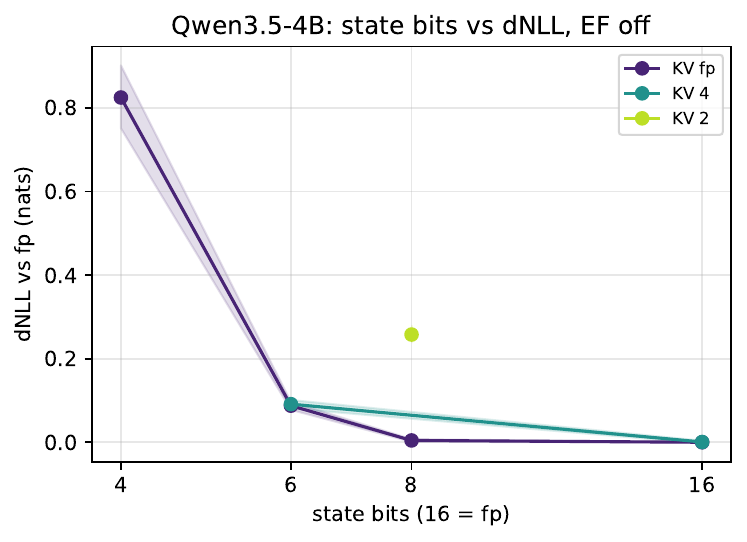}
\caption{Left: $\dnll$ of the core cells against model size, RTN, 3 seeds $\times$ 20 documents; circles
Qwen3.5, triangles Falcon-H1 (Mamba-2). Middle and right: $\dnll$ against state bits for each KV
bit-width, RTN, on 0.8B and 4B; paired bootstrap 95\% CIs.}
\label{fig:scale}
\end{figure}

%% file: tables/models.tex
\begin{tabular}{llrrrrrrrrrrrrr}
\toprule
Model & Mixer & $L$ & $L_a$ & $L_g$ & $d_{\text{model}}$ & $H_{kv}$ & $d$ & $H$ & $d_k$ & $d_v$ & $|\mathrm{KV}|/\text{tok}$ & $|S|$ (M) & $|S|$ bf16 (MB) & $T^\star$ \\
\midrule
Qwen3.5-0.8B & GDN & 24 & 6 & 18 & 1024 & 2 & 256 & 16 & 128 & 128 & 6,144 & 4.72 & 9.4 & 768 \\
Qwen3.5-2B & GDN & 24 & 6 & 18 & 2048 & 2 & 256 & 16 & 128 & 128 & 6,144 & 4.72 & 9.4 & 768 \\
Qwen3.5-4B & GDN & 32 & 8 & 24 & 2560 & 4 & 256 & 32 & 128 & 128 & 16,384 & 12.58 & 25.2 & 768 \\
Qwen3.5-9B & GDN & 32 & 8 & 24 & 4096 & 4 & 256 & 32 & 128 & 128 & 16,384 & 12.58 & 25.2 & 768 \\
Falcon-H1-1.5B-Instruct & Mamba-2 & 24 & 24 & 24 & 2048 & 2 & 128 & 48 & 64 & 256 & 12,288 & 18.87 & 37.7 & 1,536 \\
\bottomrule
\end{tabular}

%% file: tables/grid_0.8b.tex
\begin{tabular}{rrrllr}
\toprule
$b_{kv}$ & $b_s$ & Mem.\ (MB, \%) & $\Delta$NLL RTN [95\% CI] & $\Delta$NLL EF [95\% CI] & $n$ \\
\midrule
bf16 & bf16 & 15.7 (100) & $+0.000$ [+0.000, +0.000] & n/a & 3$\times$20 \\
bf16 & 8 & 11.0 (70) & $+0.015$ [+0.010, +0.020] & $+0.001$ [-0.000, +0.003] & 3$\times$20 \\
bf16 & 6 & 9.8 (62) & $+0.135$ [+0.122, +0.148] & $+0.050$ [+0.044, +0.057] & 3$\times$20 \\
bf16 & 5 & 9.2 (59) & $+0.372$ [+0.347, +0.395] & $+0.146$ [+0.133, +0.158] & 3$\times$20 \\
bf16 & 4 & 8.7 (55) & $+1.049$ [+1.001, +1.094] & $+0.524$ [+0.495, +0.553] & 3$\times$20 \\
8 & bf16 & 12.6 (80) & $-0.000$ [-0.001, +0.000] & n/a & 3$\times$20 \\
8 & 8 & 7.9 (50) & $+0.017$ [+0.012, +0.021] & $+0.002$ [-0.000, +0.003] & 3$\times$20 \\
8 & 6 & 6.7 (42) & $+0.137$ [+0.124, +0.150] & $+0.047$ [+0.041, +0.054] & 3$\times$20 \\
8 & 5 & 6.1 (39) & $+0.363$ [+0.340, +0.387] & $+0.152$ [+0.139, +0.165] & 3$\times$20 \\
8 & 4 & 5.5 (35) & $+1.045$ [+0.999, +1.092] & $+0.514$ [+0.487, +0.541] & 3$\times$20 \\
4 & bf16 & 11.0 (70) & $+0.028$ [+0.024, +0.032] & n/a & 3$\times$20 \\
4 & 8 & 6.3 (40) & $+0.043$ [+0.035, +0.050] & $+0.029$ [+0.025, +0.034] & 3$\times$20 \\
4 & 6 & 5.1 (32) & $+0.165$ [+0.151, +0.179] & $+0.076$ [+0.069, +0.084] & 3$\times$20 \\
4 & 5 & 4.5 (29) & $+0.403$ [+0.378, +0.430] & $+0.187$ [+0.172, +0.201] & 3$\times$20 \\
4 & 4 & 3.9 (25) & $+1.105$ [+1.059, +1.147] & $+0.566$ [+0.537, +0.595] & 3$\times$20 \\
3 & bf16 & 10.6 (68) & $+0.135$ [+0.125, +0.145] & n/a & 3$\times$20 \\
3 & 8 & 5.9 (38) & $+0.150$ [+0.139, +0.163] & $+0.139$ [+0.129, +0.150] & 3$\times$20 \\
3 & 6 & 4.7 (30) & $+0.289$ [+0.272, +0.308] & $+0.196$ [+0.181, +0.212] & 3$\times$20 \\
3 & 5 & 4.1 (26) & $+0.531$ [+0.502, +0.560] & $+0.320$ [+0.299, +0.340] & 3$\times$20 \\
3 & 4 & 3.5 (22) & $+1.310$ [+1.258, +1.359] & $+0.732$ [+0.700, +0.768] & 3$\times$20 \\
2 & bf16 & 10.2 (65) & $+0.439$ [+0.416, +0.464] & n/a & 3$\times$20 \\
2 & 8 & 5.5 (35) & $+0.448$ [+0.425, +0.475] & $+0.449$ [+0.424, +0.477] & 3$\times$20 \\
2 & 6 & 4.3 (28) & $+0.630$ [+0.601, +0.660] & $+0.521$ [+0.495, +0.549] & 3$\times$20 \\
2 & 5 & 3.7 (24) & $+0.913$ [+0.878, +0.950] & $+0.658$ [+0.627, +0.690] & 3$\times$20 \\
2 & 4 & 3.1 (20) & $+1.812$ [+1.759, +1.864] & $+1.177$ [+1.142, +1.215] & 3$\times$20 \\
\bottomrule
\end{tabular}

%% file: tables/ablation_0.8b.tex
\begin{tabular}{rrrccl}
\toprule
$b_{kv}$ & $b_s$ & group & symmetric & stochastic & $\Delta$NLL [95\% CI] \\
\midrule
4 & 6 & 32 & no & no & $+0.062$ [+0.053, +0.070] \\
4 & 6 & 32 & no & yes & $+0.221$ [+0.207, +0.236] \\
4 & 6 & 32 & yes & no & $+0.109$ [+0.097, +0.121] \\
4 & 6 & 32 & yes & yes & $+0.603$ [+0.574, +0.635] \\
4 & 6 & 64 & no & no & $+0.096$ [+0.086, +0.106] \\
4 & 6 & 64 & no & yes & $+0.434$ [+0.411, +0.458] \\
4 & 6 & 64 & yes & no & $+0.178$ [+0.163, +0.193] \\
4 & 6 & 64 & yes & yes & $+1.177$ [+1.132, +1.220] \\
4 & 6 & 128 & no & no & $+0.165$ [+0.151, +0.179] \\
4 & 6 & 128 & no & yes & $+0.851$ [+0.815, +0.889] \\
4 & 6 & 128 & yes & no & $+0.314$ [+0.294, +0.335] \\
4 & 6 & 128 & yes & yes & $+2.022$ [+1.950, +2.094] \\
4 & 4 & 32 & no & no & $+0.304$ [+0.287, +0.320] \\
4 & 4 & 32 & no & yes & $+2.052$ [+1.984, +2.125] \\
4 & 4 & 32 & yes & no & $+0.529$ [+0.500, +0.559] \\
4 & 4 & 32 & yes & yes & $+5.150$ [+4.997, +5.296] \\
4 & 4 & 64 & no & no & $+0.519$ [+0.494, +0.545] \\
4 & 4 & 64 & no & yes & $+3.635$ [+3.538, +3.735] \\
4 & 4 & 64 & yes & no & $+1.030$ [+0.985, +1.075] \\
4 & 4 & 64 & yes & yes & $+6.610$ [+6.443, +6.771] \\
4 & 4 & 128 & no & no & $+1.105$ [+1.059, +1.147] \\
4 & 4 & 128 & no & yes & $+5.248$ [+5.115, +5.387] \\
4 & 4 & 128 & yes & no & $+3.341$ [+3.225, +3.476] \\
4 & 4 & 128 & yes & yes & $+7.937$ [+7.750, +8.109] \\
\bottomrule
\end{tabular}

%% file: tables/grid_c4_0.8b.tex
\begin{tabular}{rrrllr}
\toprule
$b_{kv}$ & $b_s$ & Mem.\ (MB, \%) & $\Delta$NLL RTN [95\% CI] & $\Delta$NLL EF [95\% CI] & $n$ \\
\midrule
bf16 & bf16 & 15.7 (100) & $+0.000$ [+0.000, +0.000] & n/a & 3$\times$20 \\
bf16 & 8 & 11.0 (70) & $+0.025$ [+0.021, +0.030] & $+0.002$ [+0.001, +0.004] & 3$\times$20 \\
bf16 & 6 & 9.8 (62) & $+0.169$ [+0.151, +0.187] & $+0.043$ [+0.037, +0.049] & 3$\times$20 \\
bf16 & 4 & 8.7 (55) & $+1.031$ [+0.975, +1.086] & $+0.514$ [+0.484, +0.543] & 3$\times$20 \\
4 & bf16 & 11.0 (70) & $+0.023$ [+0.019, +0.027] & n/a & 3$\times$20 \\
4 & 6 & 5.1 (32) & $+0.195$ [+0.178, +0.212] & $+0.068$ [+0.062, +0.074] & 3$\times$20 \\
4 & 4 & 3.9 (25) & $+1.095$ [+1.035, +1.154] & $+0.567$ [+0.535, +0.598] & 3$\times$20 \\
2 & 8 & 5.5 (35) & $+0.425$ [+0.399, +0.452] & $+0.411$ [+0.388, +0.438] & 3$\times$20 \\
\bottomrule
\end{tabular}

%% file: tables/grid_2b.tex
\begin{tabular}{rrrllr}
\toprule
$b_{kv}$ & $b_s$ & Mem.\ (MB, \%) & $\Delta$NLL RTN [95\% CI] & $\Delta$NLL EF [95\% CI] & $n$ \\
\midrule
bf16 & bf16 & 15.7 (100) & $+0.000$ [+0.000, +0.000] & n/a & 3$\times$20 \\
bf16 & 8 & 11.0 (70) & $+0.010$ [+0.006, +0.013] & $+0.002$ [+0.000, +0.003] & 3$\times$20 \\
bf16 & 6 & 9.8 (62) & $+0.090$ [+0.080, +0.100] & $+0.024$ [+0.019, +0.029] & 3$\times$20 \\
bf16 & 4 & 8.7 (55) & $+1.808$ [+1.725, +1.896] & $+0.577$ [+0.539, +0.612] & 3$\times$20 \\
4 & bf16 & 11.0 (70) & $+0.011$ [+0.009, +0.014] & n/a & 3$\times$20 \\
4 & 6 & 5.1 (32) & $+0.106$ [+0.096, +0.117] & $+0.041$ [+0.035, +0.047] & 3$\times$20 \\
4 & 4 & 3.9 (25) & $+1.867$ [+1.787, +1.951] & $+0.612$ [+0.570, +0.650] & 3$\times$20 \\
2 & 8 & 5.5 (35) & $+0.460$ [+0.432, +0.484] & $+0.447$ [+0.417, +0.477] & 3$\times$20 \\
\bottomrule
\end{tabular}

%% file: tables/grid_4b_core.tex
\begin{tabular}{rrrllr}
\toprule
$b_{kv}$ & $b_s$ & Mem.\ (MB, \%) & $\Delta$NLL RTN [95\% CI] & $\Delta$NLL EF [95\% CI] & $n$ \\
\midrule
bf16 & bf16 & 41.9 (100) & $+0.000$ [+0.000, +0.000] & n/a & 3$\times$20 \\
bf16 & 8 & 29.4 (70) & $+0.004$ [+0.001, +0.007] & $-0.000$ [-0.002, +0.001] & 3$\times$20 \\
bf16 & 6 & 26.2 (62) & $+0.088$ [+0.078, +0.097] & $+0.037$ [+0.030, +0.045] & 3$\times$20 \\
bf16 & 4 & 23.1 (55) & $+0.825$ [+0.751, +0.902] & $+0.476$ [+0.437, +0.518] & 3$\times$20 \\
8 & 8 & 21.0 (50) & $+0.004$ [+0.001, +0.007] & $-0.001$ [-0.002, +0.001] & 3$\times$20 \\
4 & bf16 & 29.4 (70) & $+0.001$ [-0.003, +0.004] & n/a & 3$\times$20 \\
4 & 8 & 16.8 (40) & $+0.003$ [-0.001, +0.007] & $-0.001$ [-0.004, +0.003] & 3$\times$20 \\
4 & 6 & 13.6 (32) & $+0.091$ [+0.081, +0.101] & $+0.044$ [+0.036, +0.053] & 3$\times$20 \\
4 & 4 & 10.5 (25) & $+0.838$ [+0.764, +0.917] & $+0.487$ [+0.446, +0.529] & 3$\times$20 \\
2 & 8 & 14.7 (35) & $+0.258$ [+0.239, +0.279] & $+0.258$ [+0.240, +0.278] & 3$\times$20 \\
\bottomrule
\end{tabular}

%% file: tables/grid_9b.tex
\begin{tabular}{rrrllr}
\toprule
$b_{kv}$ & $b_s$ & Mem.\ (MB, \%) & $\Delta$NLL RTN [95\% CI] & $\Delta$NLL EF [95\% CI] & $n$ \\
\midrule
bf16 & bf16 & 41.9 (100) & $+0.000$ [+0.000, +0.000] & n/a & 3$\times$20 \\
bf16 & 8 & 29.4 (70) & $+0.003$ [-0.000, +0.006] & $+0.003$ [+0.001, +0.005] & 3$\times$20 \\
bf16 & 6 & 26.2 (62) & $+0.068$ [+0.056, +0.082] & $+0.058$ [+0.049, +0.066] & 3$\times$20 \\
bf16 & 4 & 23.1 (55) & $+0.283$ [+0.261, +0.303] & $+0.156$ [+0.140, +0.175] & 3$\times$20 \\
4 & bf16 & 29.4 (70) & $+0.006$ [+0.004, +0.008] & n/a & 3$\times$20 \\
4 & 6 & 13.6 (32) & $+0.076$ [+0.064, +0.090] & $+0.064$ [+0.055, +0.073] & 3$\times$20 \\
4 & 4 & 10.5 (25) & $+0.287$ [+0.265, +0.308] & $+0.164$ [+0.146, +0.182] & 3$\times$20 \\
2 & 8 & 14.7 (35) & $+0.131$ [+0.119, +0.144] & $+0.133$ [+0.121, +0.145] & 3$\times$20 \\
\bottomrule
\end{tabular}

%% file: tables/grid_falcon_1.5b.tex
\begin{tabular}{rrrllr}
\toprule
$b_{kv}$ & $b_s$ & Mem.\ (MB, \%) & $\Delta$NLL RTN [95\% CI] & $\Delta$NLL EF [95\% CI] & $n$ \\
\midrule
bf16 & bf16 & 50.3 (100) & $+0.000$ [+0.000, +0.000] & n/a & 3$\times$20 \\
bf16 & 8 & 31.5 (62) & $-0.014$ [-0.019, -0.010] & $-0.001$ [-0.002, +0.000] & 3$\times$20 \\
bf16 & 6 & 26.7 (53) & $-0.018$ [-0.024, -0.012] & $+0.002$ [-0.001, +0.004] & 3$\times$20 \\
bf16 & 4 & 22.0 (44) & $+0.004$ [-0.010, +0.018] & $-0.001$ [-0.010, +0.010] & 3$\times$20 \\
4 & bf16 & 40.9 (81) & $+0.010$ [+0.004, +0.015] & n/a & 3$\times$20 \\
4 & 6 & 17.3 (34) & $-0.009$ [-0.016, -0.002] & $+0.012$ [+0.006, +0.018] & 3$\times$20 \\
4 & 4 & 12.6 (25) & $+0.012$ [-0.003, +0.027] & $+0.010$ [-0.001, +0.019] & 3$\times$20 \\
2 & 8 & 20.4 (41) & $+0.205$ [+0.178, +0.231] & $+0.233$ [+0.207, +0.260] & 3$\times$20 \\
\bottomrule
\end{tabular}

%% file: tables/longdecode_0.8b.tex
\begin{tabular}{rrrllr}
\toprule
$b_{kv}$ & $b_s$ & Mem.\ (MB, \%) & $\Delta$NLL RTN [95\% CI] & $\Delta$NLL EF [95\% CI] & $n$ \\
\midrule
bf16 & bf16 & 15.7 (100) & $+0.000$ [+0.000, +0.000] & n/a & 3$\times$10 \\
bf16 & 8 & 11.0 (70) & $+0.051$ [+0.040, +0.062] & $+0.004$ [+0.002, +0.005] & 3$\times$10 \\
bf16 & 6 & 9.8 (62) & $+0.253$ [+0.229, +0.276] & $+0.053$ [+0.046, +0.060] & 3$\times$10 \\
bf16 & 4 & 8.7 (55) & $+1.488$ [+1.416, +1.566] & $+0.613$ [+0.579, +0.647] & 3$\times$10 \\
4 & bf16 & 11.0 (70) & $+0.027$ [+0.022, +0.031] & n/a & 3$\times$10 \\
4 & 6 & 5.1 (32) & $+0.271$ [+0.249, +0.293] & $+0.080$ [+0.071, +0.089] & 3$\times$10 \\
4 & 4 & 3.9 (25) & $+1.548$ [+1.476, +1.624] & $+0.652$ [+0.619, +0.683] & 3$\times$10 \\
\bottomrule
\end{tabular}

%% file: tables/longdecode_4b.tex
\begin{tabular}{rrrllr}
\toprule
$b_{kv}$ & $b_s$ & Mem.\ (MB, \%) & $\Delta$NLL RTN [95\% CI] & $\Delta$NLL EF [95\% CI] & $n$ \\
\midrule
bf16 & bf16 & 41.9 (100) & $+0.000$ [+0.000, +0.000] & n/a & 3$\times$10 \\
bf16 & 8 & 29.4 (70) & $+0.011$ [+0.006, +0.016] & $+0.000$ [-0.001, +0.002] & 3$\times$10 \\
bf16 & 6 & 26.2 (62) & $+0.116$ [+0.103, +0.129] & $+0.038$ [+0.030, +0.046] & 3$\times$10 \\
bf16 & 4 & 23.1 (55) & $+1.187$ [+1.076, +1.294] & $+0.562$ [+0.524, +0.601] & 3$\times$10 \\
4 & bf16 & 29.4 (70) & $+0.001$ [-0.002, +0.004] & n/a & 3$\times$10 \\
4 & 6 & 13.6 (32) & $+0.123$ [+0.111, +0.135] & $+0.039$ [+0.032, +0.046] & 3$\times$10 \\
4 & 4 & 10.5 (25) & $+1.190$ [+1.079, +1.294] & $+0.573$ [+0.534, +0.614] & 3$\times$10 \\
\bottomrule
\end{tabular}